\documentclass{article}

\usepackage{microtype}
\usepackage{graphicx}
\usepackage{subfigure}
\usepackage{booktabs} 

\usepackage{hyperref}

\usepackage[accepted]{icml2024}

\usepackage{amsmath}
\usepackage{amssymb}
\usepackage{mathtools}
\usepackage{amsthm}

\usepackage[capitalize,noabbrev]{cleveref}

\theoremstyle{plain}
\newtheorem{theorem}{Theorem}

\newtheorem{lemma}[theorem]{Lemma}
\newtheorem{corollary}[theorem]{Corollary}
\theoremstyle{definition}
\newtheorem{definition}[theorem]{Definition}
\newtheorem{example}{Example}

\theoremstyle{remark}

\usepackage{graphicx}
\usepackage{booktabs}
\usepackage{dsfont}
\usepackage{tikz}
\usetikzlibrary{positioning, arrows.meta, bending, fit, calc}
\usepackage{xcolor} 
\definecolor{darkgreen}{rgb}{0.0, 0.5, 0.0} 
\usepackage{bm}
\usepackage{amsmath,amssymb,amsthm}
\usepackage{mathtools}
\usepackage{pifont}

\newlength{\symbolwidth}
\definecolor{cb-green-sea}  {RGB}{  0, 146, 146}
\definecolor{cb-burgundy}   {RGB}{146,   0,   0}
\newcommand{\cmark}{\makebox[\symbolwidth]{\textcolor{cb-green-sea}{\ding{51}}}}
\newcommand{\xmark}{\makebox[\symbolwidth]{\textcolor{cb-burgundy}{\ding{55}}}}

\DeclareMathOperator*{\argmin}{arg\,min}

\newcommand{\wtilde}[1]{\stackrel{\sim}{\smash{{#1}}\rule{0pt}{1.1ex}}}

\usepackage[textsize=tiny]{todonotes}

\icmltitlerunning{Generalizing Orthogonalization for Models with Non-Linearities}

\begin{document}

\twocolumn[
\icmltitle{Generalizing Orthogonalization for Models with Non-Linearities}



\icmlsetsymbol{equal}{*}

\begin{icmlauthorlist}
\icmlauthor{David R\"ugamer}{lmu,mcml}
\icmlauthor{Chris Kolb}{lmu,mcml}
\icmlauthor{Tobias Weber}{lmu,mcml}
\icmlauthor{Lucas Kook}{vie}
\icmlauthor{Thomas Nagler}{lmu,mcml}
\end{icmlauthorlist}

\icmlaffiliation{lmu}{Department of Statistics, LMU Munich, Munich, Germany}
\icmlaffiliation{mcml}{Munich Center for Machine Learning (MCML), Munich, Germany}
\icmlaffiliation{vie}{Institute for Statistics and Mathematics, Vienna University of Economics and Business, Vienna, Austria}

\icmlcorrespondingauthor{David R\"ugamer}{david@stat.uni-muenchen.de}
\icmlkeywords{Machine Learning, ICML}

\vskip 0.3in
]



\printAffiliationsAndNotice{}  


\begin{abstract}
The complexity of black-box algorithms can lead to various challenges, including the introduction of biases. These biases present immediate risks in the algorithms’ application. It was, for instance, shown that neural networks can deduce racial information solely from a patient's X-ray scan, a task beyond the capability of medical experts. If this fact is not known to the medical expert, automatic decision-making based on this algorithm could lead to prescribing a treatment (purely) based on racial information. While current methodologies allow for the ``orthogonalization'' or ``normalization'' of neural networks with respect to such information, existing approaches are grounded in linear models. Our paper advances the discourse by introducing corrections for non-linearities such as ReLU activations. Our approach also encompasses scalar and tensor-valued predictions, facilitating its integration into neural network architectures. Through extensive experiments, we validate our method's effectiveness in safeguarding sensitive data in generalized linear models, normalizing convolutional neural networks for metadata, and rectifying pre-existing embeddings for undesired attributes.
\end{abstract}

\section{Introduction}

In the burgeoning landscape of artificial intelligence and deep learning, black-box algorithms have become a centerpiece for driving advances in many fields of application. These powerful and often inscrutable models offer impressive predictive capabilities, but their complexity also gives rise to challenges that cannot be overlooked. One of the most urgent contemporary challenges is the correction of unwanted behaviors in these algorithms. In particular, the presence of biases in model predictions and learned representations can lead to unintended consequences that transcend the technological sphere, impacting societal norms and ethical considerations.
For example, \citet{glocker2022risk} and \citet{weber2023unreading} realized that predictions of convolutional networks trained for chest X-ray pathology classification are heavily affected by implicitly encoded racial information, leading to potentially inaccurate or unfair medical assessments.

In this paper, we stress the necessity of addressing these challenges and propose several solutions, focusing on what has been termed ``orthogonalization'' or ``normalization'' of neural networks in the literature \citep[see, e.g.,][]{lu2021metadata, pho}. These approaches can be used to adjust predictions, prevent unequal treatment of population subgroups, protect from unintentionally revealing sensitive information, or ensure the interpretability of black-box models. 
\begin{table*}[!t]
    \centering
        \caption{Estimated influence (coefficients) of sex, age, and race on the model's predictions when using self-supervised embeddings of the MIMIC-CXR dataset \cite{sellergren2022simplified} on the label \textit{Pleural Effusion}. The first row shows uncorrected values, the second row the results after classical orthogonalization, and the third row after applying our generalized orthogonalization. p-values in brackets indicate whether these coefficients represent a significant influence (p-values smaller than 0.05 show a \xmark-sign, otherwise a \cmark-sign).}
    \label{tab:mimic}
    \vskip 0.1in
    \resizebox{0.95\textwidth}{!}{
    \begin{tabular}{rrrrr}
     & Sex (male) & Age & Race (Black) & Race (White)  \\ \hline 
   Without correction  & 0.064 ({9.3e-9})\,\xmark & 2.237 ({$<$ 2e-16})\,\xmark  & -0.483 ({$<$ 2e-16})\,\xmark & -0.050 (0.0743) \,\cmark \\
   Classical orthogonalization $\mathcal{C}^l$ & 0.014 (0.192)\,\cmark & 0.453 ({$<$ 2e-16})\,\xmark  & -0.117 ({$<$ 2e-4})\,\xmark & -0.015 (0.603)\,\cmark \\
   Generalized orthogonalization $\mathcal{C}^h$ & 0.001 (0.994)\,\cmark & -0.011 (0.689)\,\cmark  & -0.001 (0.753)\,\cmark & -0.000 (0.904)\,\cmark
    \end{tabular}
    }
\end{table*}
\subsection{Related Literature} \label{sec:rellit}

\paragraph{Classical Orthogonalization} In contrast to research investigating network weight dependencies \citep[see, e.g.,][]{huang2020controllable}, the orthogonalization discussed in this work is usually motivated as a normalization operation or correction method. \citet{lu2021metadata}, e.g., proposed the so-called metadata normalization network, which applies a layer-wise orthogonalization to remove metadata information from layers. Similarly, \citet{he2019learning} study an orthogonalization procedure to encourage fair learning by removing protected information from feature representations. To circumvent training instabilities caused by dependence on the batch size, \citet{vento2022penalty} propose to cast the metadata normalization as a bi-level problem optimized using a penalty approach. \citet{kaiser2022uncertaintyaware} use the same idea to incorporate fairness in automated decision-making systems for labor market data. A related but different problem 
is addressed by \citet{rugamer2023semi}, who propose an orthogonalization cell to preserve effect identifiability and thereby interpretability. This idea was then also adapted to other model classes \citep{baumann2021deep, kopper2021semi}. Further, orthogonalization can also be used as a debiasing technique \citep[e.g., in graph neural networks,][]{9381348}. Similar to our work, orthogonalization can be applied as part of a neural network \citep[e.g.,][]{lu2021metadata} or post-model fitting \citep{pho}. The latter approach also has a link to double and debiased machine learning in econometrics \citep[see, e.g.,][]{chernozhukov2018double}, where a related correction is performed to allow for inference statements about a treatment variable after removing (potentially non-linear) nuisance effects.

\paragraph{Fair Machine Learning} Another related strand of literature is fair machine learning \citep{chen2022scalable}. As we discuss later in our work, methods to achieve fairness are closely linked to orthogonalization and vice versa. A wide range of notions and approaches exists to operationalize the concept of fairness \citep[see, e.g.,][for literature reviews]{mehrabi2021survey,pessach2022review}. Proposed methods are applied at different steps, e.g., in data pre-processing \citep{calmon2017optimized} or post-processing \citep{hardt2016equality,sattigeri2022fair,xu2022controlling}, and have been derived for different model classes, such as generalized linear models \citep[GLMs;][]{do2022fair} or kernel learning \citep{perez2017fair}. These approaches are usually implemented as constrained \citep[see, e.g.,][]{pmlr-v80-komiyama18a, zafar2019fairness} or regularized optimization problems \citep[e.g.,][]{scutari2022achieving,do2022fair} and require making a trade-off between model performance and the amount of achieved fairness. In contrast, our approach forces the model predictions to be uncorrelated with the protected information, thereby ensuring that they can not be inferred to any degree from the predictions using a linear model. While this potentially comes at the cost of a loss in model performance, a conservative correction routine that errs on the side of caution is required in many risk-averse domains such as medical applications with sensitive patient information \citep{glocker2023algorithmic}, or when the goal is to achieve model identifiability \citep{pho}. A limitation of presented fairness solutions is that the sensitive information is made explicit in the model under the assumption of a linear effect on the outcome \citep[see, e.g.,][]{scutari2022achieving}.

Despite the versatility of orthogonalization in practice, most existing methods share the same working principle which assumes \underline{linearity} in both the model that needs to be corrected and the correction function. In many real-world situations, however, models include non-linearities that render the classical orthogonalization unable to fully correct the model (cf.~Table~\ref{tab:mimic} with details in Section~\ref{sec:mimic}). 

\subsection{Our Contribution}

In this paper, we extend the orthogonalization operation to non-linearity in both the model that needs to be corrected and the correction function. We derive a correction routine that works for a variety of non-linear models, including GLMs and neural networks. Moreover, we show how to extend our approach to arbitrary tensor-valued predictions. This allows our method to be flexibly inserted anywhere in common neural network architectures and thereby further generalizes the idea of normalization. Our experimental results demonstrate the efficacy of our approach, showing how 1) sensitive information can be successfully protected in GLMs, 2) neural networks with non-linear activation functions can be normalized for metadata during training, and 3) orthogonalization effectively corrects pre-trained embeddings for unwanted attributes.

\section{Orthogonalization} \label{sec:ortho}

Given an input $\bm{X} \in \mathbb{R}^{n \times p}, p \leq n$ with full column rank, the linear predictor in a parametric learning model (or the pre-activation of a fully-connected neural network layer) is given by $\bm{X}\bm{\beta}$, where $\bm{\beta} \in \mathbb{R}^p$ denotes the model's or layer's weights. In a linear model, we can write the least squares predictions $\hat{\bm{y}}$ as a function of the inputs $\bm{X}$ and targets $\bm{y}$: 
\begin{equation*}
\hat{\bm{y}} = \bm{X}\hat{\bm{\beta}} = \mathcal{P}_{\bm{X}} \bm{y},
\end{equation*}
where $\mathcal{P}_{\bm{X}}=\bm{X}(\bm{X}^{\top} \bm{X})^{-1}\bm{X}^{\top}$ is the projection matrix projecting $\bm{y}$ onto $\text{span}(\bm{X})$, the space spanned by the columns of $\bm{X}$. 
Instead of projecting onto $\text{span}(\bm{X})$, we can also project onto the orthogonal complement $\text{span}(\bm{X})^\bot$ using $\mathcal{P}_{\bm{X}}^\bot \bm{y}=(\bm{I}_n - \mathcal{P}_{\bm{X}})\bm{y}$,  where $\bm{I}_n$ is the identity matrix. Hence, when left-multiplying a term by $\mathcal{P}_{\bm{X}}^\bot$, we \emph{orthogonalize} this term (w.r.t.~features $\bm{X}$).
\begin{figure*}[!ht]
    \centering
    \includegraphics[width=0.9\textwidth]{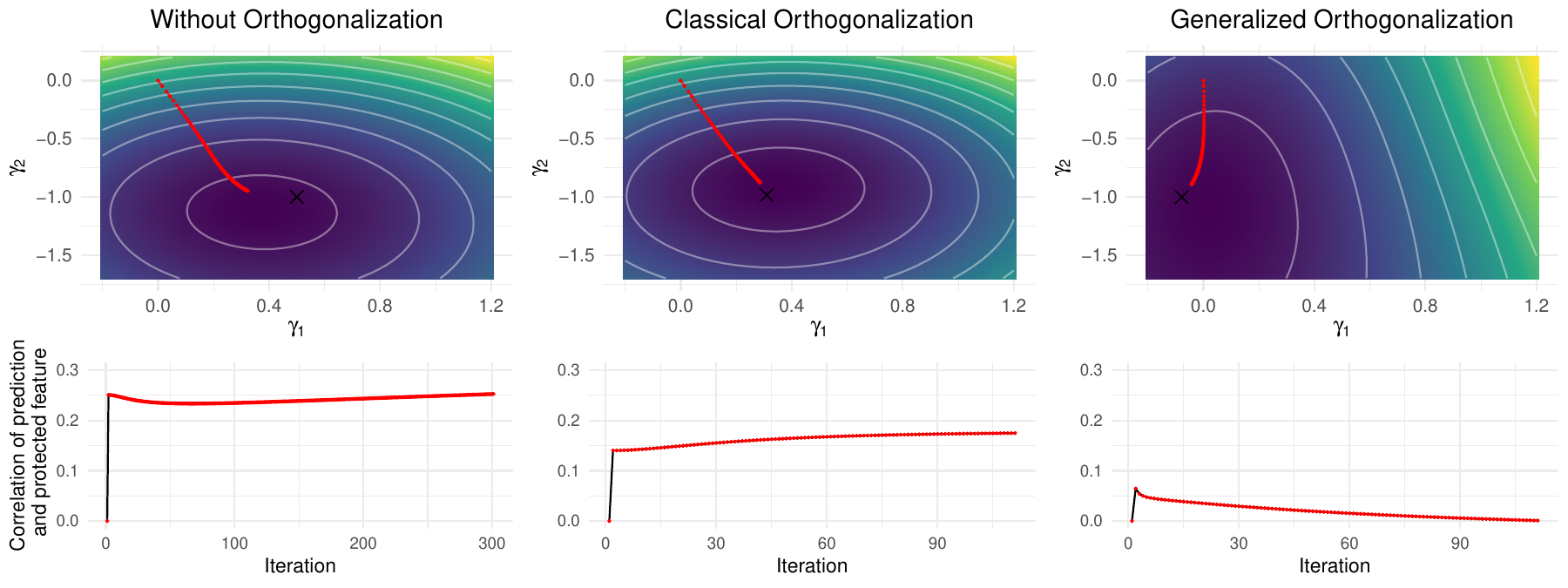}
    \caption{Exemplary optimization process for a fixed number of iterations (converging towards the black crosses) for logistic regression with features $z_1,z_2$, weights $\gamma_1,\gamma_2$, and one protected feature $x_1$ correlated with $z_1$. The upper row shows the loss surface and optimization path for the three different methods (columns) a small step size. The bottom row shows the correlation between the model's prediction and the protected feature along the optimization path, where only the generalized orthogonalization yields predictions uncorrelated with the sensitive information.}
    \label{fig:demo_plot}
\end{figure*}
Ultimately --- whether the motivation is debiasing, protecting information, or identifiability --- the universal recipe for most approaches discussed in Section~\ref{sec:rellit} is to use the described orthogonalization operation. 
Given protected features $\bm{X}$, which are not supposed to influence the modeling process, and a model that uses a different set of features $\bm{Z} \in \mathbb{R}^{n\times q}, q \geq p$ for prediction, we can ``correct'' the model (or, in this case equivalently, the features $\bm{Z}$) by fitting the model with $\bm{Z}^c := \mathcal{P}_{\bm{X}}^\bot \bm{Z}$ instead of $\bm{Z}$.
This is equivalent to removing all linear $\bm{X}$-information in $\bm{Z}$. Note that $\bm{Z}$ might also contain features from $\bm{X}$, but the correction is supposed to work for unobserved relationships between $\bm{Z}$ and $\bm{X}$. We call this correction function $\mathcal{C}^l$ in the following.

To see that $\mathcal{C}^l$ in fact removes the effect of protected features $\bm{X}$ from $\bm{Z}$ in the prediction, we can check whether the corrected predictions $\hat{\bm{y}}^c = \bm{Z}^c \hat{\bm{\gamma}}^c$, based on the orthogonalized features $\bm{Z}^c = \mathcal{P}_{\bm{X}}^\bot \bm{Z}$ and a corrected weight vector $\hat{\bm{\gamma}}^c \in\mathbb{R}^q$, can be linearly explained by the features $\bm{X}$ to any degree. This can be done by regressing $\hat{\bm{y}}^c$ on $\bm{X}$. If the correction is successful, the debiased effect 
$$\hat{\bm{\beta}}^c := \argmin_{\bm{\beta}^c} \Vert \hat{\bm{y}}^c - \bm{X\beta^c}\Vert_2^2 
$$
of $\bm{X}$ in the linear model with outcome $\hat{\bm{y}}^c$ and features $\bm{X}$ should be $\hat{\bm{\beta}}^c \equiv \bm{0}$. Plugging in the terms of the ordinary least squares solution for both $\hat{\bm{\beta}}^c$ and $\hat{\bm{\gamma}}$, we get 
\begin{equation*}
    \begin{split}
        \hat{\bm{\beta}}^c &= (\bm{X}^\top\bm{X})^{-1}\bm{X}^\top \hat{\bm{y}}^c = (\bm{X}^\top\bm{X})^{-1}\bm{X}^\top \bm{Z}^c \hat{\bm{\gamma}} \\
        &= (\bm{X}^\top\bm{X})^{-1}\bm{X}^\top \mathcal{P}_{\bm{X}}^\bot \bm{Z} (\bm{Z}^\top \mathcal{P}_{\bm{X}}^\bot \bm{Z})^{-1} \bm{Z}^\top \mathcal{P}_{\bm{X}}^\bot \bm{y},
    \end{split}
\end{equation*}
which yields $\bm{0}$ as $\bm{X}^\top \mathcal{P}_{\bm{X}}^\bot = \bm{0}_{p \times n}$. This confirms that we have successfully removed any \underline{linear} effect of $\bm{X}$ from the predictions $\hat{\bm{y}}^c$ and corrected features $\bm{Z}^c$. So when $\bm{Z}$ is, e.g., a pre-trained embedding, using $\bm{Z}^c$ instead of $\bm{Z}$ in a downstream task removes the partial influence of protected features $\bm{X}$ contained in $\bm{Z}$, as we first project the original embedding onto a space that is orthogonal to the space spanned by the protected features $\bm{X}$.

\section{Orthogonalization for Models with Non-linearities}

While the classical orthogonalization allows correcting for metadata of images~\citep{lu2021metadata} or identifying contributions in additive predictors \citep{pho}, all previous methods fail to provide a valid correction in the case of a non-linear, element-wise transformation $h$ applied to $\bm{Z\gamma}$, i.e., $\hat{\bm{y}} = h(\bm{Z\gamma})$, such as in GLMs or most neural network layers (see Figure~\ref{fig:demo_plot} for an example). In the following, we present an extension to previously proposed orthogonalization approaches that also encompass models with non-linear transformations of the linear predictor, thereby accounting for various important use cases. For simplicity, we motivate our approach using a GLM, but also discuss its embedding in neural architectures in Sections~\ref{sec:nlproj}--\ref{sec:nlmn}. Before deriving the theoretical details, we provide an instructive example in the following.

\begin{example} \label{ex:1}
    Following \citet{weber2023unreading}, assume we are given an embedding $\bm{Z} \in \mathbb{R}^{n \times q}$ from a medical imaging task and want to ensure that by using this embedding, we do not share any protected patient information encoded in $\bm{X}$. Assume that the embedding contains information from $\bm{X}$ \citep[which represents a very real problem; see][]{glocker2022risk}. When predicting the patients' disease status using the embedding $\bm{Z}$ in a GLM with 
    $\hat{\bm{y}} = h(\bm{Z}\hat{\bm{\gamma}})$, where $h$ is the sigmoid function, we risk making decisions implicitly based on protected features. Instead, we have to come up with a corrected GLM routine such that the corrected non-linear model predictions $\hat{\bm{y}}^c$ do not utilize or leak patient information $\bm{X}$.
    \end{example}

\subsection{Types of Orthogonalizations} \label{sec:easycase}

We first start by examining the properties of the original orthogonalization $\mathcal{C}^l$ when applied to models with non-linear activation functions. In particular, it is worth noting that using $\bm{Z}^c = \mathcal{P}_X^\bot \bm{Z}$ from Section~\ref{sec:ortho} \emph{can} be used in a GLM when the goal is to orthogonalize the model's additive predictor $\bm{\eta} = \bm{Z\gamma}$. In this case, a corrected version $\bm{\eta}^c = \bm{Z}^c{\bm{\gamma}}$ will be orthogonal to the column space of $\bm{X}$ for any $\bm{\gamma}$ and hence, \emph{before the non-linear transformation} $h$, ``free'' of any (linear) effect of protected features $\bm{X}$. In other words, regressing $\bm{\eta}^c$ on $\bm{X}$ will result in effects $\hat{\bm{\beta}}^c = \bm{0}$ as derived for the linear case in Section~\ref{sec:ortho}. 
In contrast, when transforming $\bm{\eta}^c$ with $h$, the model space becomes non-linear and 
regressing $\hat{\bm{y}} = h(\bm{\eta}^c)$ on $\bm{X}$ will generally not yield effects $\hat{\bm{\beta}}^c = \bm{0}$, i.e., the transformed predictor $h(\bm{\eta}^c)$ still potentially contains $\bm{X}$-information (cf.~Appendix~\ref{app:simplified} for an illustrative explanation). As illustrated in Example~\ref{ex:1}, such non-linear transformations are ubiquitous in machine learning and employed in most classification tasks.

When using a non-linear activation function $h$ in the prediction of $\bm{y}$ based on $\bm{Z}$, a natural choice is to also use the same non-linear transformation to check the model's prediction for any $\bm{X}$ influence. To better clarify the different types of models involved in an orthogonalization and to unify previous endeavors, we define the following terms for the workflow described in Figure~\ref{fig:workflow}:
\begin{figure}[!t]
    \centering
    \resizebox{0.9\columnwidth}{!}{
\begin{tikzpicture}[
  block/.style={rectangle, draw, rounded corners, minimum height=1.5cm, minimum width=2cm}, 
  arrow/.style={-Latex},
  dashedarrow/.style={-Latex, dashed, line width=0.25pt}, 
  font=\Large
]

\node[block] (Mp) {$\bm{\mathcal{M}}^p$};
\node[block, below=1cm of Mp] (Me) {$\bm{\mathcal{M}}^e$}; 
\node[left=2cm of Mp.west] (input1) {$\bm{Z}, \bm{y}$};
\node[left=2cm of Me.west] (input2) {$\bm{X}, \bm{\hat{y}}^{\textcolor{cb-green-sea}{\textbf{c}}}$};
\node[right=2cm of Mp.east] (output1) {$\bm{\hat{y}}^{\textcolor{cb-green-sea}{\textbf{c}}}$};
\node[right=2cm of Me.east] (output2) {$\bm{\hat{\beta}}^{\textcolor{cb-green-sea}{\textbf{c}}} \overset{?}{=} \bm{0}$};
\node[draw, dashed, cb-green-sea, fit=(Mp), inner sep=0.1cm, line width=1.25pt, rounded corners] (box) {};
\node[above right=0.01cm and 0.01cm of box.north east, cb-green-sea] (ch) {$\mathcal{C}^h$};

\draw[arrow] (input1) -- (Mp);
\draw[arrow] (Mp) -- (output1);
\draw[arrow] (input2) -- (Me);
\draw[arrow] (Me) -- (output2);
\draw[arrow] (output1) to[out=-90,in=90] (input2);

\end{tikzpicture}
}
\caption{Workflow of generating predictions via model $\mathcal{M}^p$ and checking the influence of $\bm{X}$ on the resulting predictions $\hat{\bm{y}}$ using $\mathcal{M}^e$. Green parts indicate the orthogonalization applied to $\mathcal{M}^p$.}
\label{fig:workflow}
\end{figure}
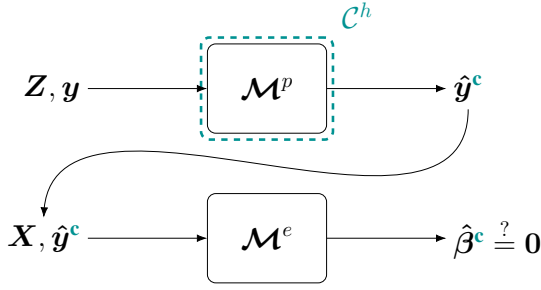
%
\begin{definition} \label{def1}
    Given features $\bm{Z}\in\mathbb{R}^{n \times q}$, a monotonic activation function $h$, loss function $\ell:\mathbb{R}^n \times \mathbb{R}^n \to \mathbb{R}$, and protected features stored in a full column rank matrix $\bm{X} \in \mathbb{R}^{n\times p}$, we define
    \begin{itemize}
        \item the \textbf{prediction model} $\mathcal{M}^p: \mathbb{R}^{n\times q} \to \mathbb{R}^n,\, \bm{Z} \mapsto \hat{\bm{y}} = h(\bm{Z}\hat{\bm{\gamma}})$, the model using $\bm{Z}$ to generate predictions $\hat{\bm{y}}$ for an outcome of interest $\bm{y}$, with parameters $\hat{\bm{\gamma}}$ learned by minimizing $\ell$;
        \item the \textbf{correction routine} $\mathcal{C}^h:$ A routine correcting $\mathcal{M}^p$ to produce corrected predictions $\hat{\bm{y}}^c$;
        \item the \textbf{evaluation model} $\mathcal{M}^e: \mathbb{R}^{n\times p} \times \mathbb{R}^n \to \mathbb{R}^p,$ $(\bm{X},\hat{\bm{y}}^c) \mapsto \hat{\bm{\beta}}^c$, the model minimizing $\ell(\hat{\bm{y}}^c,  h(\bm{X}\hat{\bm{\beta}}^c))$ to check whether the influence of $\bm{X}$ on the corrected predictions $\hat{\bm{y}}^c$ from $\mathcal{M}^p$ has been successfully removed by $\mathcal{C}^h$.
    \end{itemize}
\end{definition}

The previous definition assumes that the activation function $h$ is the same for all three components $\mathcal{M}^p$, $\mathcal{M}^e$, and $\mathcal{C}^h$. 


Before introducing our new orthogonalization algorithm when the activation function $h$ is used, we more formally define what we mean by ``correcting for the influence of $\bm{X}$'': 

\begin{definition} \label{def2}
Given the setup and models in Definition~\ref{def1}, we say
    the model predictions $\hat{\bm{y}}$ of $\mathcal{M}^p$ are \textbf{corrected} (or \textbf{orthogonalized}) if  $\mathcal{M}^e$ yields coefficient values $\hat{\bm{\beta}}^c = \bm{0}$. $\hat{\bm{y}}^c$ are then called \textbf{corrected predictions}.
\end{definition}

While these definitions imply a similar goal as pursued in fair machine learning, our approach uses a slightly different measure of ``unbiasedness'' that coincides with the previously introduced normalization \citep{lu2021metadata} as well as the orthogonalization for identifiability \citep{rugamer2023semi}. In contrast, in fair machine learning, the level of unfairness is, e.g., defined as the proportion of the variance of $\hat{\bm{y}}$ that can be explained by $\bm{X}$ \citep[see, e.g.,][or further explanations in Appendix~\ref{app:fairness}]{scutari2022achieving}. Using this notion, a model adhering to Definition~\ref{def2} is perfectly fair, but the opposite is not always true. This is because fairness is achieved using a criterion different from the one in Definition~\ref{def2}. In addition, these approaches often do not fully enforce fairness to preserve other model properties, in most cases only use the $L_2$ loss for $\ell$ and assume a linear relationship between $\bm{X}$ and $\bm{y}$.

\subsection{Orthogonalization for GLM-type Routines} \label{sec:nlproj}

We first extend the classical orthogonalization of Section~\ref{sec:ortho} motivated by linear models to GLMs. This model class is described by a distributional assumption inducing a loss function $\ell$ defined by the corresponding negative log-likelihood and strictly monotone activation function $h$ (equal to the so-called inverse link function in GLMs). The implied loss function comprises many common machine learning setups: binary classification using sigmoid activation, multi-class classification using a softmax-type activation, or, e.g., a count regression using the $\exp$-activation function. When applying Definition~\ref{def2} to these GLMs using their canonical link function, we say that the correction of a model $\mathcal{M}^p$ yielding $\hat{\bm{y}}^c$  is successful if the evaluation model  $\mathcal{M}^e$ defined by the minimizer
\begin{equation} \label{eq:glm}
    \hat{\bm{\beta}}^c = \argmin_{\bm{\beta}^c} \ell(\hat{\bm{y}}^c, h(\bm{X}\bm{\beta}^c))
\end{equation}
is a null model, i.e., $\hat{\bm{\beta}}^c \equiv \bm{0}$. Note that we here assume a model without intercept for better readability\footnote{The intercept does not necessarily have to be equal to zero, as it does not use any information from the protected features}. While the solution of \eqref{eq:glm} is known not to have an analytical solution in general \citep [see, e.g.,][]{nelder1972generalized}, the GLM poses a convex optimization problem and can be solved efficiently with iteratively reweighted least squares or Newton-type procedures \citep{gill2019generalized}. 

Importantly, the iterates of these Newton-type optimization procedures can be represented as
    $\hat{\bm{\beta}}^{[t]} = (\bm{X}^\top\bm{\Upsilon}^{[t]}\bm{X})^{-1} \bm{X}^\top \bm{\Upsilon}^{[t]} \bm{r}^{[t-1]},\,t\in\mathbb{N}$,
where the weights $\bm{\Upsilon}^{[t]} \in \mathbb{R}^{n \times n}$ are given by a diagonal matrix and $\bm{r}^{[t]} \in \mathbb{R}^{n}$ is a working response (see Appendix~\ref{app:glm} for details). As both $\bm{\Upsilon}^{[t]}$ and $\bm{r}^{[t-1]}$ are fixed after fitting the evaluation model, we can make use of the approximate linearity of both $\mathcal{M}^p$ and $\mathcal{M}^e$ at convergence to derive our generalized orthogonalization. A helpful intermediate result is the following:
\begin{lemma} \label{lemma:1}
Given any GLM model $\mathcal{M}^p$ with predictions $\hat{\bm{y}}$, and GLM model $\mathcal{M}^e$ with features $\bm{X}$,
\begin{equation}
    \hat{\bm{y}}^c = \mathcal{P}_{\bm{X}}^\bot \hat{\bm{y}} + h(0) \bm{1}_n    
\end{equation}
defines corrected predictions yielding $\hat{\bm{\beta}}^c = \bm{0}$.
\end{lemma}
\cref{lemma:1} is an interesting and important finding to construct a correction, stating that despite the non-linearity of $\mathcal{M}^e$, it is sufficient to (linearly) orthogonalize predictions of $\mathcal{M}^p$ using the orthogonal projection $\mathcal{P}_{\bm{X}}^\bot$. The derivation and proof can be found in Appendix~\ref{app:deriv_lemma}. However, as we discuss in the following paragraph, it is not straightforward to relate this finding to a corrected GLM model $\mathcal{M}^p$. 

\paragraph{Deriving corrected coefficients}

In order to derive a corrected GLM routine for corrected coefficients $\hat{\bm{\gamma}}^c$ that are consistent with $\hat{\bm{y}}^c$, a na\"ive approach is to simply invert their definition $\hat{\bm{y}}^c = h(\bm{Z}\hat{\bm{\gamma}}^c)$, yielding
\begin{equation*} 
    \hat{\bm{\gamma}}^c = \bm{Z}^\dagger h^{-1}(\hat{\bm{y}}^c) = \bm{Z}^\dagger h^{-1}( \mathcal{P}_{\bm{X}}^\bot \hat{\bm{y}}),
\end{equation*}
where $\bm{Z}^\dagger = (\bm{Z}^\top\bm{Z})^{-1}\bm{Z}^\top$ is the pseudo-inverse of $\bm{Z}$. However, as $\text{dom}(\mathcal{P}_{\bm{X}}^\bot \hat{\bm{y}})$ is not necessarily a subset of $\text{dom}(\hat{\bm{y}})$, $h^{-1}$ might not be defined on this new domain. In classification tasks, for example, this could lead to predicted probabilities $\hat{\bm{y}}^c \notin (0,1)$ 
for which the inverse sigmoid function is not defined. Since this problem will occur whenever non-linear functions $h$ or $h^{-1}$ are employed, we instead have to adapt the GLM optimization routine as described in the following. 
\begin{corollary} \label{cor:constrOptim}
Given activation function $h$, prediction model $\mathcal{M}^p$ and evaluation model $\mathcal{M}^e$ with features $\bm{X}$, the solution $\bm{\gamma}^c$ to the optimization problem
\begin{equation} \label{eq:constrOptim}
  \begin{gathered}
    \argmin_{\bm{\gamma}} \ell(\bm{y},h(\bm{Z\gamma}))\\ 
    s.t.~||(\bm{X}-n^{-1} \bm{1}_n^{\phantom{\top}}\bm{1}_n^\top \bm{X})^\top h(\bm{Z\gamma})||_2^2 = 0
\end{gathered}
\end{equation}
produces corrected coefficients satisfying $\hat{\bm{\beta}}^c = \bm{0}$.
\end{corollary} 
The mathematical details and derivations are given in the Appendix~\ref{app:theo1}. While various routines exist to solve \eqref{eq:constrOptim}, the problem turns out to be hard in practice, in particular in applications that are concerned with fairness, such as the ones addressed in Section~\ref{sec:expGLM}. We found the \emph{modified differential multiplier method} \citep{platt1987constrained} to be very effective and describe its details in Appendix~\ref{app:theo1}. 
Further, note that in the case of an identity activation function $h$, the derived optimization problem in \cref{cor:constrOptim} is also directly solved by using $\mathcal{C}^h = \mathcal{C}^l$.

\subsection{Orthogonalization with Piece-wise Linear Activations}

While the previous orthogonalization can be used in many different situations, intermediate network layers are often specified with piece-wise linear activation functions not comprised by the GLM framework, most notably the ReLU function. As for canonical activation functions in GLMs, the ReLU activation 
    $\text{ReLU}(\bm{X\beta}) = \bm{X\beta} \circ \mathds{1}(\bm{X\beta} > 0)$ 
is applied element-wise, with $\circ$ denoting the Hadamard product and $\mathds{1}(\cdot)$ an indicator function individually evaluated for all entries of the vector. While at first glance the definition of the ReLU activation might suggest that $\text{ReLU}(\bm{X\beta}) \in \text{span}(\bm{X})$ and hence $\mathcal{C}^l$ from Section~\ref{sec:easycase} is sufficient for successful orthogonalization, this is not the case as can be verified with a simple counterexample.
\begin{example}
    Assume $\bm{X} = (1,-1)^\top \in \mathbb{R}^2$ and $\beta = 1$. Then $\text{ReLU}(\bm{X}\beta)=(1,0) \notin \text{span}(\bm{X})$.
\end{example}
Although $\text{ReLU}(\bm{X\beta})$ does not necessarily lie in $\text{span}(\bm{X})$, we can derive the following result
\begin{theorem} \label{thm:piece}
Given a prediction model $\mathcal{M}^p$ with ReLU activation, and an evaluation model $\mathcal{M}^e$ with ReLU activation and $L_2$ loss, a $\mathcal{C}^h$-orthogonalization of the form $\bm{Z}^c = \mathcal{P}_{\bm{X}}^\bot \bm{Z}$, will yield $\hat{\bm{\beta}}^c = \bm{0}$.
\end{theorem}
The proof of Theorem~\ref{thm:piece} is given in the Appendix. 

\subsection{Orthogonalization for Tensor Predictions} \label{sec:nlmn}

We now show that our concept of orthogonalization is not restricted to regression models, but can also be used to remove 
information from learned representations in layers of neural networks. 
Such a correction should be able to deal with prediction models (or layers) with tensor-shaped predictions (e.g., present in early layers of convolutional networks) and we show how to adapt previous results accordingly. 
We first define the linear evaluation model $\mathcal{M}^e$ with $L_2$-loss $\ell$ as the solution of the following tensor-on-vector regression:
\begin{equation} \label{eq:tosr}
    \hat{\mathfrak{B}}^c = \argmin_{\mathfrak{B}^c} \ell(\hat{\mathfrak{Y}}^c, \bm{X} \times_1 \mathfrak{B}^c),
\end{equation}
where the outcome tensor $\hat{\mathfrak{Y}}^c \in \mathbb{R}^{n \times d_1 \times \cdots \times d_R}$ is the corrected version of an (intermediate) non-linear prediction $\hat{\mathfrak{Y}}$, $\bm{X} \in \mathbb{R}^{n\times p}$ is a feature matrix as before, $\mathfrak{B}^c \in \mathbb{R}^{p \times d_1 \times \cdots \times d_R}$ a tensor-valued regression coefficient, and $\times_1$ denotes the $1$-mode product. For a successful correction $\mathcal{C}^h$ from which we obtain $\hat{\mathfrak{Y}}^c$, it must hold $\hat{\mathfrak{B}}^c \equiv \bm{0}$ in \eqref{eq:tosr}, which can be achieved by projecting $\text{vec}(\hat{\mathfrak{Y}})$ onto $\text{span}(\bm{X})^\bot$ as before. Indeed, using a vectorized reformulation, we can prove that this yields corrected predictions (see Appendix~\ref{app:tensor} for details):
\begin{corollary} \label{cor:tensor}
Given the set-up around the evaluation model defined by \eqref{eq:tosr}, the corrected predictions satisfying $\hat{\mathfrak{B}}^c \equiv \bm{0}$ are given by
\begin{equation} \label{eq:tensor}
\hat{\mathfrak{Y}}^c = \mathcal{P}_{\bm{X}}^\bot \times_1 \hat{\mathfrak{Y}}.
\end{equation}     
\end{corollary}
While the result in \eqref{eq:tensor} can be extended to GLM-type non-linearities, and hence an orthogonalization as the one in Section~\ref{sec:nlproj} can be derived, we here focus on the ReLU activation function, arguably the pervasive type of non-linearity in intermediate DNN layers. 
Hence, we consider an evaluation model $\mathcal{M}^e$ with ReLU activation defined by
\begin{equation} \label{eq:tosr2}
    \hat{\mathfrak{B}}^c = \argmin_{\mathfrak{B}^c} \ell(\hat{\mathfrak{Y}}^c, \text{ReLU}(\bm{X} \times_1 \mathfrak{B}^c)),
\end{equation}
and a prediction model $\mathcal{M}^p$ yielding predictions $\hat{\mathfrak{Y}} = \text{ReLU}(\bar{\mathfrak{Y}})$ with pre-activation $\bar{\mathfrak{Y}}$. In contrast to previous sections, we here focus on correcting the predictions and not the effects of $\mathcal{M}^p$, as predictions of an intermediate layer in a large neural network are not necessarily formed by a linear operation. We obtain the following result:
\begin{corollary} \label{cor:relu2}
    Assuming the evaluation model \eqref{eq:tosr2}, the corrected pre-activations $\bar{\mathfrak{Y}}^c$ yielding corrected predictions $\hat{\mathfrak{Y}}^c$ following ReLU activation are given by $\bar{\mathfrak{Y}}^c = \mathcal{P}_{\bm{X}}^\bot \times_1 \bar{\mathfrak{Y}}$. 
\end{corollary}
See Appendix~\ref{app:theo3} for a proof. In other words, by multiplying the first dimension of a non-activated output of some layer with $\mathcal{P}_{\bm{X}}^\bot$ from the left before applying the ReLU activation, we obtain corrected predictions in the next layer.

\section{Numerical Experiments}

In Section~\ref{sec:expGLM}, we first check whether sensitive information can be successfully protected in GLMs with our approach, using both synthetic and real-world datasets. We then show in various settings how our method corrects learned representations for potential biases (Section~\ref{sec:expEMB}), and finally investigate its effectiveness as an online version applied during training (Section~\ref{sec:expOO}). The details of all experiments are provided in Appendix~\ref{app:model}. In all experiments, categorical information in the protected features is included in $\bm{Z}$ as one-hot encoded features with the first column removed (as we don't want to orthogonalize or penalize the model for a constant intercept term).

\subsection{Generalized Linear Model} \label{sec:expGLM}

In the following, we focus on real-world applications but provide synthetic data examples in Appendix~\ref{app:syn}, confirming that $\mathcal{C}^h$ works as intended for GLMs in a variety of settings.

\subsubsection{Adult Income Data}

Using the adult income data also investigated in \citet{xu2022controlling} to analyze algorithm fairness, we check the efficacy of our approach for protected features sex and race. We therefore first fit our prediction model, a GLM with features age, work class, education, marital status, relationship, and working hours per week, to predict the person's income binarized into $\leq 50$k and $>50$k (as provided by the original dataset) and then correct for the protected features. 

\textbf{Results}\, We find (Table~\ref{tab:adult_transposed}, first row) that the income predictions are significantly influenced if the individual is white and/or male with a multiplicative increase in the odds-ratio of approximately exp(0.522) = 1.68  and exp(1.066) = 2.90  for earning more than 50k for white individuals and males, respectively. Similar results are inferred from \citet{xu2022controlling}'s approach. 
When applying our framework 
and adjusting for sex and race, we can see that all effects of these protected features are estimated to be close to zero and are non-significant. 
%
%
\begin{table}[!ht]
    \centering
    \caption{Estimated coefficients from $\mathcal{M}^e$ with corresponding p-values in brackets when not correcting the model (first column), when using \citet{xu2022controlling} (second column), and when correcting with $\mathcal{C}^h$ (last column).  P-values smaller than 0.05 show a \xmark-sign, otherwise a $\cmark$-sign.}
    \label{tab:adult_transposed}
    \resizebox{1.0\columnwidth}{!}{
    \begin{tabular}{lrrr}
     & w/o correction & \citet{xu2022controlling} & using $\mathcal{C}^h$ \\ \hline
    Sex (male) & 1.066 ({$<$ 2e-16})\,\xmark & 1.255 ({$<$ 2e-16})\,\xmark & -0.000 (0.998)\,\cmark \\
    Race (Asian) & 0.263 (0.227)\,\cmark & 0.859 ({2e-4})\,\xmark & 0.011 (0.954)\,\cmark \\
    Race (Black) & -0.094 (0.577)\,\cmark & 0.191 (0.319)\,\cmark & 0.008 (0.957)\,\cmark \\
    Race (Other) & -0.165 (0.594)\,\cmark & -0.131 (0.706)\,\cmark & -0.131 (0.609)\,\cmark \\
    Race (White) & 0.522 ({0.001})\,\xmark & 0.928 ({4e-7})\,\xmark & 0.005 (0.969)\,\cmark \\
    \end{tabular}
    }
\end{table}
\begin{figure*}[!ht]
    \centering
    \includegraphics[width=0.9\textwidth]{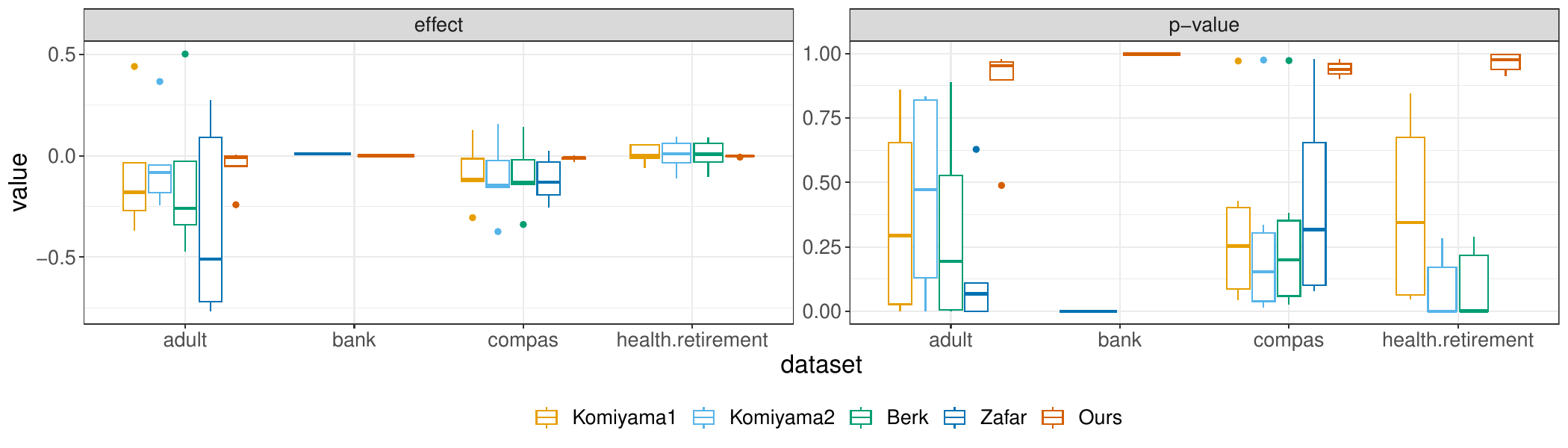}
    \caption{Comparison of orthogonalization properties of fairness methods in comparison with our approach. Ideally, methods should yield effects of zero (in the left plot) and large p-values (in the right plot). Missing boxes indicate that the method did not converge.}
    \label{fig:fair}
\end{figure*}
\subsubsection{Fairness Benchmark} \label{sec:fairnessbench}

While fairness methods target a slightly different goal than orthogonalization, 
it might still be interesting to see whether our criterion for orthogonalization (cf.~Definition~\ref{def2}) can also be met by approaches suggested in the fairness literature. We therefore run a comprehensive analysis on fairness datasets with non-Gaussian outcomes and compare the results of various fairness methods with our orthogonalization. In the case of GLMs, closely related methods are fair (logistic / Poisson) regression with different optimization strategies. \citet{pmlr-v80-komiyama18a} bounds the proportion of explained variance by protected features (we refer to this as \emph{Komiyama1}). Another variant by the same authors first regresses the fitted values against the protected features (referred to as \emph{Komiyama2}). An alternative is the approach by \citet{berk2017convex} enforcing individual fairness by penalizing pairs of observations that have different outcomes for the same protected features (referred to as \emph{Berk}). A fourth approach we compare against is the one of \citet{zafar2019fairness} (referred to as \emph{Zafar}) enforcing fairness by bounding the covariance of protected features and other predictors. We run these and our method on the benchmark datasets \texttt{adult}, \texttt{bank}, \texttt{compas}, and \texttt{health.retirement} provided in \citet{scutari2023fairml} containing various protected features such as race, sex, or marriage. 

\textbf{Results}\, The results are depicted in Figure~\ref{fig:fair} showing that all fairness approaches show significant feature effects with effect strengths often different from 0. In contrast, our approach adheres to the orthogonality requirement from Definition~\ref{def2}, yielding effects close to 0 with no significant impact. As stated earlier, this does not imply unfairness of the other approaches, but 
demonstrates that fairness approaches cannot be simply used as a substitute to achieve orthogonality.
As orthogonality might potentially decrease model performance, Table~\ref{tab:performance_fair} additionally compares the model performances with and without the correction.
\begin{table}[htbp]
  \centering
  \small
  \caption{Test set prediction performances with/out correction.}
    \begin{tabular}{lcc}
    Dataset (Metric) & \multicolumn{1}{l}{w/o Corr.} & \multicolumn{1}{l}{w/ Corr.} \\
    \midrule
    adult (Accuracy) & 0.854 & 0.833 \\
    bank (Accuracy) & 0.899 & 0.887 \\
    compas (Accuracy) & 0.739 & 0.724 \\
    health (Root mean squared error) & 0.750 & 0.789 \\
    \end{tabular}%
  \label{tab:performance_fair}%
\end{table}%
Since most of the datasets contain the protected features also in $\bm{Z}$, i.e., as features for the prediction model, the prediction performance with orthogonalization will naturally decrease (if not, then the protected features would likely not bias the model and raise fairness issues in the first place). However, as can be seen in Table~\ref{tab:performance_fair}, the decrease in performance is often negligible with a maximum of 2\% decrease in accuracy for compas.

\subsection{Orthogonalization of Learned Representations} \label{sec:expEMB}

\subsubsection{Comparing $\mathcal{C}^l$ and $\mathcal{C}^h$ on Image Embeddings} \label{sec:mimic}

Going back to our instructive Example 1, we use self-supervised embeddings for the MIMIC Chest X-Ray dataset \citep{johnson2019mimic, sellergren2022simplified} known to have encoded race, sex and age \citep{glocker2022risk}, and try to remove this implicit information using orthogonalization. Following \citet{glocker2022risk}, we first validate that meta-information such as race, in fact, is influencing predictions in a downstream task. As $\mathcal{M}^p$ we use a GLM to predict whether the patient suffers from pleural effusion using only the embedding as features. We then check whether a $\mathcal{C}^l$-orthogonalization is sufficient to remove this information, but in contrast to the Adult data application, check on the level of probabilities where $\mathcal{C}^l$ should fail. Finally, we apply $\mathcal{C}^h$-orthogonalization to obtain the corrected predictions.

\textbf{Results}\, Table~\ref{tab:mimic} (page 2) summarizes the results, confirming our hypotheses and demonstrating that our approach effectively removes the meta-information from embeddings. In this case, removing the features even results in a slight improvement in prediction performance with a 0.02 increase in AUC.

\subsubsection{Post-hoc Orthogonalization for Neural Representations}

As another demonstration, we analyze the race and age bias in a learned neural representation of faces from the UTKFace dataset \citep{zhifei2017cvpr}, where previous literature suggests, among other biases, a misalignment for black women \citep[see, e.g.,][]{krishnan2020understanding}.  We first train a ResNet50 \citep{he2016deep} on the provided images to classify the sex of the depicted person. We then check the influence of race and age on the predictions using a logistic regression. In order to correct for biases, we extract the weights in the ResNet's penultimate layer and apply the $\mathcal{C}^h$-orthogonalization by fitting a corrected GLM (following \cref{cor:constrOptim}) with learned representations as features and sex as the binary outcome. Finally, we again check the influence of race and age on the model's predictions.

\textbf{Results}\, Results are summarized in~Table~\ref{tab:faces_transposed}, again depicting coefficients and p-values, and confirm the effectiveness of our approach in blinding the network for certain features (here age and race). The orthogonalization in this case comes at the cost of an AUC reduction from 0.831 (w/o correction) to 0.826 (with $\mathcal{C}^h$), which seems to be a reasonable trade-off to correct for age and race effects.

%
\begin{table}[!ht]
    \centering
    \caption{Estimated coefficients from $\mathcal{M}^e$ with corresponding p-values in brackets when not correcting the model (first column) and when correcting the model for age and race (second column). Significant influences are highlighted in bold for an $\alpha$-level of 0.05.}
    \label{tab:faces_transposed}
    \resizebox{\columnwidth}{!}{
    \begin{tabular}{lrr}
     & w/o correction & $\mathcal{C}^h$ orthogonalization \\ \hline 
    Age & -0.017 (\,{$<$ 2e-16})\,\xmark & -0.000 (0.976)\,\cmark \\
    Race (Black) & -0.352 ({8.68e-10})\,\xmark & 0.009 (0.872)\,\cmark \\
    Race (Indian) & -0.620 (\,{$<$ 2e-16})\,\xmark & 0.008 (0.894)\,\cmark \\
    Race (Other) & -0.315 ({4.05e-05})\,\xmark & -0.002 (0.979)\,\cmark \\
    Race (White) & -0.377 ({5.65e-13})\,\xmark & -0.008 (0.872)\,\cmark \\
    \end{tabular}
    }
\end{table}

\subsubsection{Word Embeddings}

As a last example, we demonstrate that our approach is not limited to tabular and image data. As, e.g., discussed in \citet{bolukbasi2016man}, text embeddings, as well, can have biases encoded, and addressing this property seems of great relevance given the rise of GPT(-like) models which have also been shown to not always be neutral in their generation \citep[see, e.g.,][]{ray2023chatgpt}. To show that the proposed method also works for the correction of text embeddings in a non-linear model, we use the movies review dataset \citep{maas-EtAl:2011:ACL-HLT2011}, which we enhance with the gender information of the protagonist and the movie director. Using an LSTM model and a learned text embedding $\bm{Z}$ (see Appendix~\ref{app:model} for details), we try to predict the number of reviews for each movie (i.e., a Poisson-distributed outcome). We do this once without correction and once by correcting the embeddings $\bm{Z}$ of size $100 \times 100$ of every movie description following the results from \cref{cor:relu2} using the available gender information as $\bm{X}$.

\textbf{Results}\, When not correcting the model's prediction, we obtain p-values $<$2e-16 for both gender effects, suggesting a significant influence of gender on the predictions $\hat{\bm{y}}$. After correction, p-values are 0.993 and 1.0, implying a successful embedding correction with only a small increase in root mean squared error of $1\%$ compared to the model without correction.

\subsection{Online Orthogonalization} \label{sec:expOO}

Finally, we test how our approach performs when applied in an online fashion by iteratively optimizing a small CNN and orthogonalizing the network with respect to some fixed features $\bm{X}$ as proposed in the Metadata Normalization approach \citep{lu2021metadata}. In other words, we perform the proposed orthogonalization during training. Inspired by Clever Hans predictors \citep{lapuschkin2019unmasking} and orthogonalization in distribution shift situations \citep{chen2023project}, we colorize the MNIST data by reshaping the grayscale image to an RGB image where one of the three channels is filled with the original grayscale pixel values and the remaining two channels filled with zeros, i.e., set to black. The binary prediction task is to classify a subset of the MNIST dataset reduced to digits $0$ and $9$. For the train set, we intentionally create only red images of $0$s by assigning the pixel values to the first channel. For the images of $9$s we randomly assign either the second (green) or third (blue) channel. In the counterfactual test set, images of both classes are colored randomly in green or blue, but not red. As a result, the color red is predictive for the train but not the test set. We then use the color information ``red'', denoted as the binary feature $\bm{X}$, in the correction model to normalize the CNN's predictions. In other words, we train a CNN on RGB images but blind the network w.r.t.~the color red (or in fact any) color information by orthogonalizing the first convolutional layer's latent features w.r.t.~$\bm{X}$.
\begin{figure}[!ht]
    \centering
    \includegraphics[width = 0.95\columnwidth]{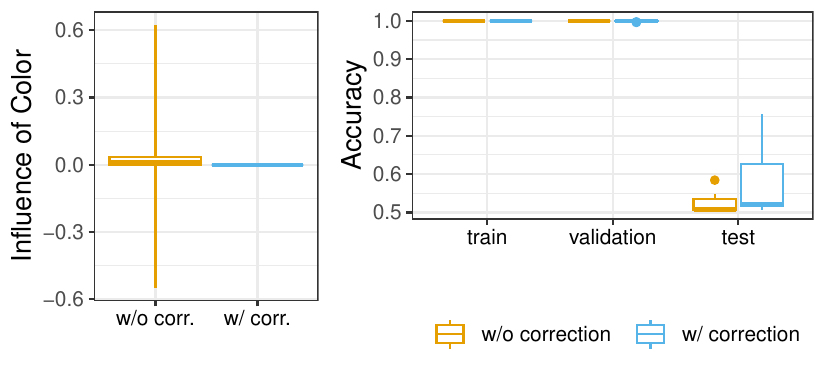}
    \caption{Estimated influence of the color red on the convolutional layer's predictions (left plot) and train-, validation- and test-accuracy of the two models (with/out correction).}
    \label{fig:mnist}
\end{figure}

\textbf{Results}\, Figure~\ref{fig:mnist} summarizes the results of the experiment showing that our method correcting for the color red in fact yields no influence of the metadata information on the hidden layer's weights (left plot) and further illustrates how the model without correction is fooled by the training/validation data by focusing on the information of the color channels instead of the pixel values. This leads to a poor test performance of not much more than 50\% accuracy. The corrected model shows a better test performance as it learned to focus on the image content and not the channel information.

\section{Discussion}
\begin{figure*}[t]
    \centering
    \includegraphics[width=0.9\textwidth]{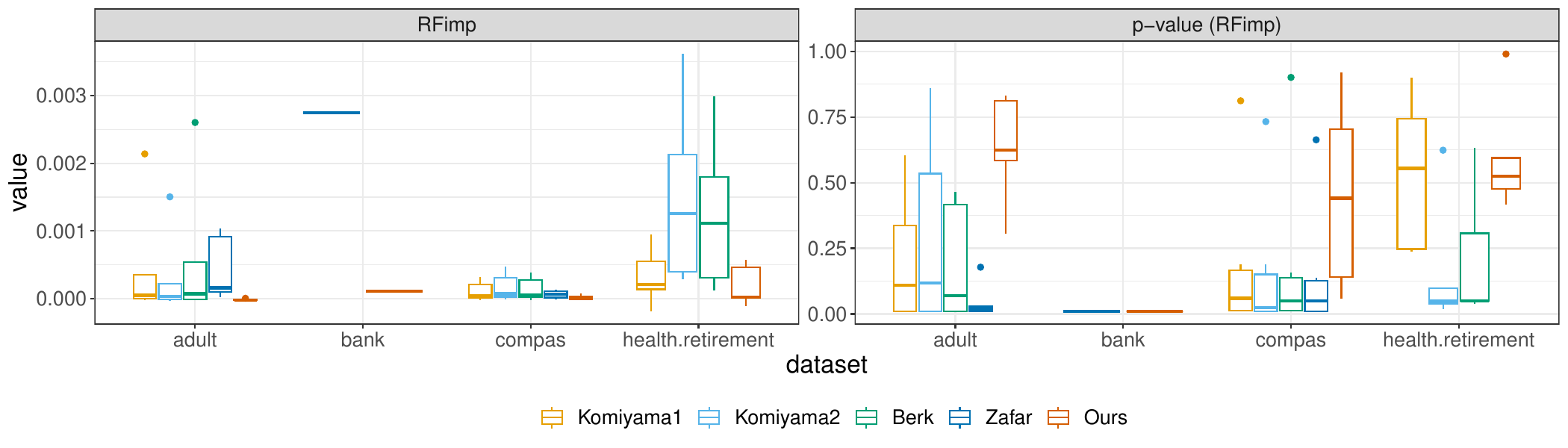}
    \caption{Comparison of orthogonalization properties of fairness methods in comparison with our approach when extending the notion of orthogonality beyond linear feature effects. Ideally, methods should yield random forest feature importance (RFimp) of zero (in the left plot) and large p-values (in the right plot). Missing boxes indicate that the method did not converge.}
    \label{fig:randfor}
\end{figure*}
In this work, we introduced a novel correction routine that extends the concept of orthogonalization to non-linear and tensor-valued models. Our experimental results confirm that our approach is effective in correcting sensitive data in linear models and neural networks, as well as rectifying pre-existing embeddings for undesired attributes.

\textbf{Caveat}\, We want to emphasize that the given examples and results in this paper are limited by the representativeness of the datasets used and should not be the sole basis for decision-making in applications involving medical imaging or text applications. Furthermore, as ``protection of information'' can be defined in various ways, our presented methods do not offer guarantees of absolute protection for sensitive information in every situation. In particular, note that orthogonalization does not imply stochastic independence of the involved random variables, which means that some form of residual dependence could persist in theory. 

\textbf{Limitations and future research}\, The proposed algorithms require the number of observations $n$ to be larger than the number of features $p$ (or intermediate layer predictions in a network), and that there are more features $q$ in the prediction model than in the evaluation model ($p$). While this is usually the case, extensions to applications for $n < p$ or $q < p$ are a challenging yet interesting future research direction.

The proposed orthogonalization also requires knowledge of features $\bm{Z}$ related to protected information. Performing such an operation to remove implicit biases is another interesting approach brought up by an anonymous reviewer but would require simultaneously learning the protected feature and being able to evaluate whether these present potential biases.

\section{Outlook}
A possible extension suggested by one of the anonymous reviewers is to make the class of evaluation models more flexible, allowing it to go beyond linear effects of protected features on the additive predictor scale, i.e., fitting an evaluation model $h(g(X))$ where $g$ is, e.g., a non-linear function. This would yield a more general notion of orthogonalization than presented in this paper but is also more challenging to analyze theoretically. In order to check how robust existing approaches and ours are w.r.t.~non-linear evaluation models, we run the fairness benchmark again and use a random forest as an evaluation model to explain the prediction model’s output using protected information. We then extract the importance of each feature together with a p-value for the importance values using a permutation test. Results are depicted in Figure~\ref{fig:randfor}, showing a very similar pattern to Figure~\ref{fig:fair}. Interestingly, our method seems to work well in terms of this new orthogonalization notion, only yielding significant feature importances for the bank dataset. Results for all other methods suggest that protected features play a significant role in the random forest’s explanation of the prediction model’s output.
%

\section*{Impact Statement}
This paper presents work whose goal is to advance the field of machine learning. There are many potential societal consequences of our work, none of which we feel must be specifically highlighted here.

\bibliography{references}
\bibliographystyle{icml2024}

\newpage
\appendix
\onecolumn

\section{Simplified Problem Statement} \label{app:simplified}

\begin{figure}[!ht]
    \centering
\begin{tikzpicture}[scale=1, transform shape]
  
  \draw[thick,->] (-1,0) -- (3,0) node[right] {x};
  \draw[thick,->] (0,-1) -- (0,3) node[above] {y};
  
  \draw[thick, blue,->] (0,0) -- (1,0) node[below right] {$\mathbf{x} = [1, 0]$};
  \draw[thick, blue,->] (0,0) -- (0,1) node[above left, yshift=2pt] {$\mathbf{y} = [0, 1]$};
  
  \draw[thick, red,->] (0,0) -- (2.718,1) node[above] {$h_{\exp}(\mathbf{x}) = [e, 1]$};
  \draw[thick, red,->] (0,0) -- (1,2.718) node[right] {$h_{\exp}(\mathbf{y}) = [1, e]$};
  
  \pgfmathsetmacro{\sigmoidx}{exp(1)/(exp(1)+1)}
  \pgfmathsetmacro{\sigmoidy}{1/2}
  \draw[thick, darkgreen,->] (0,0) -- (\sigmoidx,\sigmoidy) node[right] {$h_{\sigma}(\mathbf{x}) = [\frac{e}{e+1}, \frac{1}{2}]$};
  \draw[thick, darkgreen,->] (0,0) -- (\sigmoidy,\sigmoidx) node[left, xshift=-12pt,yshift=-8pt] {$h_{\sigma}(\mathbf{y}) = [\frac{1}{2}, \frac{e}{e+1}]$};
  
\end{tikzpicture}
    \caption{Graphical depiction of orthogonality without transformation (blue vectors) and after transformation with $h_{\exp}$ (red) or the sigmoid function $h_\sigma$ (green).}
    \label{fig:vector}
\end{figure}
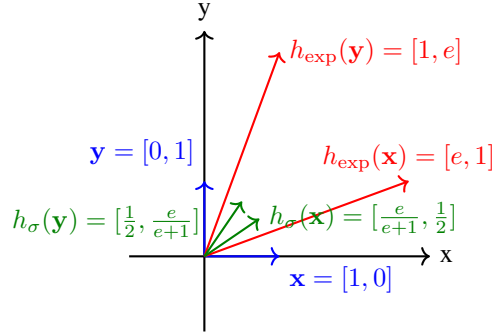

To illustrate our problem, consider two vectors $x=(0,1)$ and $y=(1,0)$, which are orthogonal as $x^\top y = 0$. However, after transformation $h(x)$ and $h(y)$ are not orthogonal anymore for non-linear transformations $h$ as depicted in Figure~\ref{fig:vector}.

\section{Connection to Fair Machine Learning} \label{app:fairness}

When comparing the original orthogonalization approach in \citet{lu2021metadata,rugamer2023semi} and \citet{pmlr-v80-komiyama18a}, both approaches use corrected features
\begin{equation}
    \bm{Z}^c = \mathcal{P}_{\bm{X}}^\bot \bm{Z}
\end{equation}
in their model equation. The main difference is that fair machine learning approaches incorporate the protected features into the analysis as well, i.e., estimate the model
\begin{equation} \label{eq:fair}
    \bm{y} = \bm{X}\bm{\beta} + \bm{Z}^c\bm{\gamma}^c + \bm{\varepsilon}
\end{equation}
with residual term $\bm{\varepsilon}$, whereas the prediction model $\mathcal{M}^p$ in the orthogonalization approach does not necessarily include the features $\bm{X}$ in the model. If $\bm{X}$ is incorporated and without further constraints on the model, the approach of \citet{pmlr-v80-komiyama18a} and the orthogonalization would lead to the same effect estimates \citep[e.g., when using orthogonalization for identifiability as in][]{pho}. Fair machine learning, however, offers a trade-off between model performance and (un-)fairness by regularizing the effects $\bm{\beta}$ of $\bm{X}$ in \eqref{eq:fair} using an explained variance formulation:
\begin{equation}
\min_{\bm{\beta},\bm{\gamma}^c} \mathbb{E}[(\bm{y}-\hat{\bm{y}})^2] \quad s.t.~\frac{\text{Var}(\bm{X\beta})}{\text{Var}(\hat{\bm{y}})} \leq r
\end{equation}
with predictions $\hat{\bm{y}} = \bm{X}\hat{\bm{\beta}} + \bm{Z}^c\hat{\bm{\gamma}}^c$ and some constraint $r\in[0,1]$. This approach and reformulations, such as the one of \citet{scutari2022achieving} as a ridge-penalized optimization problem, hence do not provide the full unpenalized effects $\bm{\beta}$ as obtained by the orthogonalization.

Apart from the difference between the two approaches induced by the regularization, the generalization of fair machine learning approaches to models with activation functions differs further. In particular, as, e.g, suggested by \citet{scutari2022achieving} the orthogonal predictors $\bm{Z}^c$ are only used on the pre-activation scale, i.e., the fair model uses $\bm{Z}^c$ in the GLM by incorporating it only once in the additive predictor: $\mathbb{E}(\bm{y}|\bm{X},\bm{Z}) = h(\bm{X}\bm{\beta} + \bm{Z}^c\bm{\gamma}^c)$. As discussed in the main part of our paper and in the previous section, this does not imply orthogonality of $h(\bm{X\beta})$ and $h(\bm{Z\gamma})$ as required in our work.

A slightly different approach is used by \citet{xu2022controlling}  who correct a classifier trained with all variables by another classifier trained only on the protected variables. Their approach, however, does not debias representations but only finds directions for predictions that are orthogonal.

\section{Other Notions of Orthogonality}

In the following, we briefly describe connections to other notions of orthogonality that are used in different fields of machine learning research.

\subsection{Domain Generalization}

The objective of the domain generalization (DG) problem is to train models to be robust against (or orthogonal to) domain (or covariate) shifts \citep{ganin2016domain,shankar2018generalizing}. While our orthogonalization approach and DG have different goals, viewing orthogonalization through the lens of DG is an interesting idea brought up by an anonymous reviewer. In particular, distributionally robust optimization \citep[DRO;][]{sagawa2019distributionally} aims to ensure that predictive accuracy is robust w.r.t.~changes in the distribution of the sensitive features. Our orthogonalization approach, in contrast, aims to ensure that the sensitive features do not affect the predictions (in the sense of a zero coefficient in the evaluation model). Note, however, that DRO does not provide any guarantees regarding our goal. 

\subsection{Orthogonality of Predictions and Embeddings}

Another interesting question brought up by an anonymous reviewer is how our notion of orthogonality relates to the following two concepts:
\begin{itemize}
    \item \textbf{Concept 1}: requiring that the predictions do not change when the sensitive attributes are changed;
    \item \textbf{Concept 2}: requiring that the embeddings are free from the influence of sensitive information.
\end{itemize}
Concept 2 is a special case of our proposed notion if the model is linear. Concept 1 can also coincide with our definition if sensitive attributes are part of the prediction model. In our setting, however, this is not necessarily the case. Other (fairness) approaches aligning with concept 1 assume knowledge of the causal relationship between sensitive and non-sensitive attributes, which we also do not require. 

\section{Background on GLMs} \label{app:glm}

We briefly introduce several quantities used in the estimation of GLMs, but refer to \citet{wood2017generalized} for more detailed explanations and derivations. Let $g = h^{-1}$ be the link function. The weights $\bm\Upsilon$ are defined as
\begin{equation} \label{eq:GLMweight}
    \bm\Upsilon = \text{diag}(\alpha(\mu_i)/\{g'(\mu_i)^2 V(\mu_i)\})
\end{equation}
where $\mu_i = h(\bm{x}_i^\top \bm{\beta})$ and
\begin{equation}
   \alpha(\mu_i) = 1 + (y_i - \mu_i)\{V'(\mu_i)/V(\mu_i) + g''(\mu_i)/g'(\mu_i)\}
\end{equation}
and
\begin{equation}
    V(\mu_i) = \text{Var}(y)/\phi
\end{equation}
with $\phi$ denoting the distribution's scale parameter. The working response $\bm{r}$ is defined as
\begin{equation}
    \bm{r} = \bm{G}(\bm{y}-\bm{\mu})
\end{equation}
where $\bm{\mu} = (\mu_1,\ldots,\mu_n)^\top$ and
\begin{equation} \label{eq:GLMgradient}
     \bm{G} = \text{diag}(g'(\mu_i) / \alpha(\mu_i)).
\end{equation}
In particular, note that in the case of using \emph{Fisher weights}, i.e., the expected Hessian matrix, $\alpha(\mu_i)=1$ and we have the following simplification: 
\begin{equation} \label{eq:constfac}
    \bm{G}^{-1}\bm{\Upsilon}^{-1/2} = \text{diag}(V(\mu_i)^{1/2}),
\end{equation}
i.e., a function of $\mu_i$ only. We can use this result in later derivations. 


\subsection{GLM in Detail} \label{app:glmwdetail}

The above derivation for GLMs can be made more explicit for our purposes. As noted before, using the Fisher weights implies $\alpha(\mu_i) = 1$. As $\mu_i = h(x_i^\top \beta) \equiv h(0)$ in case of a successful correction, this implies 
\begin{equation}
    \bm{G} = \text{diag}(g'(\mu_i)) = (h^{-1})'(0) \cdot \bm{I}.
\end{equation}
In other words, multiplying with $\bm{G}$ only scales all observations by a constant. Similarly, \eqref{eq:constfac} reduces to a constant multiplicative factor of $V(h(0)^{1/2})$. 

\section{Proofs and Derivations}

\subsection{Proof of \cref{lemma:1}} \label{app:deriv_lemma}

To prove \cref{lemma:1}, we use the final iterate representation of weights $\hat{\bm{\beta}}^c$ in a GLM and note that the evaluation model (first without intercept) must yield
\begin{equation} \label{eq:betahat2}
    \hat{\bm{\beta}}^c = (\bm{X}^\top\bm{\Upsilon}\bm{X})^{-1} \bm{X}^\top \bm{\Upsilon} \bm{r} = (\bm{X}^\top\bm{\Upsilon}\bm{X})^{-1} \bm{X}^\top \bm{\Upsilon} \bm{G} (\hat{\bm{y}}^c - \bm{\mu}) \overset{!}{=} \bm{0}.
\end{equation}
Consequently, it must also hold $\bm{\mu} = h(\bm{X} \hat{\bm{\beta}}^c) = h(\bm{0})$. The equality in \eqref{eq:betahat2} yields two possible solutions: either $\hat{\bm{y}}^c = \bm{\mu}$, which is undesirable as it implies that we cannot predict anything after removing the influence of $\bm{X}$, or we can construct $\hat{\bm{y}}^c$ so that when first subtracting $\bm{\mu}$ and then projecting it onto the space spanned by the columns of $\wtilde{\bm{X}} = \bm{\Upsilon}^{1/2} \bm{X}$, we obtain the zero vector (see details below). Solving for $\hat{\bm{y}}^c$, we get the following result.
\begin{lemma} \label{lemma1}
Consider a prediction model $\mathcal{M}^p$ with activation function $h$ using data $\bm{Z}$ to produce predictions $\hat{\bm{y}}$. Given weights $\bm{\Upsilon}$ and protected features $\bm{X}$, corrected predictions satisfying $\hat{\bm{\beta}}^c=\bm{0}$ are given by
\begin{equation} \label{eq:nlo}
    \hat{\bm{y}}^c = \bm{G}^{-1}\bm{\Upsilon}^{-1/2} \mathcal{P}_{\widetilde{\bm{X}}}^\bot \hat{\bm{y}} + h(\bm{0}).
\end{equation}
\end{lemma}
We can easily confirm that this yields a desired solution by plugging in \eqref{eq:nlo} into \eqref{eq:betahat2}. 
Further note that $\bm{\mu} = h(0) \cdot \bm{1}_n$, i.e., shifts by a constant $c_2 := h(0)$ and the multiplication from the left with $\bm{G}^{-1}\bm{\Upsilon}^{-1/2}$ does only scale by a constant factor $c_1 := V(h(0))^{1/2}$ (see Section~\ref{app:glmwdetail}) but not rotate $\mathcal{P}_{\widetilde{\bm{X}}}^\bot \hat{\bm{y}}$, i.e.,
\begin{equation} \label{eq:nlo_w_const}
        \hat{\bm{y}}^c = \bm{G}^{-1}\bm{\Upsilon}^{-1/2} \mathcal{P}_{\widetilde{\bm{X}}}^\bot \hat{\bm{y}} + \bm{\mu}
        \overset{(\ref{eq:constfac})}{=} V(h(0))^{1/2}  \mathcal{P}_{\widetilde{\bm{X}}}^\bot \hat{\bm{y}} + h(\bm{0}) = c_1 \mathcal{P}_{\widetilde{\bm{X}}}^\bot \hat{\bm{y}} + c_2 \bm{1}_n
\end{equation}
with constants $c_1, c_2$. Any multiplicative constant does not change the result.
We can therefore further simplify the previous expression from \eqref{eq:nlo} to
\begin{equation} \label{eq:reqzerohat}
    \hat{\bm{y}}^c = \mathcal{P}_{\bm{X}}^\bot \hat{\bm{y}} + c_2 \bm{1}_n.
\end{equation}
This is due to the fact that $\mathcal{P}_{\widetilde{\bm{X}}}^\bot$ is the projection matrix of $\widetilde{\bm{X}} = \bm{\Upsilon}^{1/2} \bm{X}$ with $\bm{\Upsilon}^{1/2}$ being $\bm\Upsilon^{1/2} = \text{diag}(1/(g'(\mu_i) \sqrt{V(\mu_i)})) \equiv c_3 \cdot \bm{I}$ for some constant $c_3$ as $\bm{\mu} = h(\bm{0}) = const. \cdot \bm{1}_n$, and hence both $g'(\mu_i)$ and $V(\mu_i)$ are constant. As a consequence, we have $\mathcal{P}_{\widetilde{\bm{X}}} = c_3 \bm{X} (c_3^2 \bm{X}^\top\bm{X})^{-1} \bm{X} c_3 = \mathcal{P}_{\bm{X}}$. 
For a similar reason, we can ignore $c_1$ as $(\bm{X}^\top\bm{X})^{-1} \bm{X}^\top c_1 \mathcal{P}_{\bm{X}}^\bot \hat{\bm{y}} = c_1(\bm{X}^\top\bm{X})^{-1} \bm{X}^\top \mathcal{P}_{\bm{X}}^\bot \hat{\bm{y}} = \bm{0}$.

\paragraph{Model with Intercept} Similar results can be derived for a model with intercept. First, note that $\bm{\mu} = h(\beta_0 \bm{1}_n)$. Hence, $c_1$ and $c_2$ are fixed constants $c_2 = h(\beta_0)$ and $c_1 = V(h(\beta_0))^{1/2}$. Hence, this yields again \eqref{eq:reqzerohat}, now just with different constants $c_1$ and $c_2$. 
In an intercept-only GLM with canonical link function, the intercept can be computed using only the response values, allowing to derive the closed-form solution $\hat{\beta}_0 = h^{-1}(\bar{\hat{y}})$ with $\bar{\hat{y}} = n^{-1} \sum_i \hat{y}_i$.
As the intercept in the evaluation model is of no particular interest, one can alternatively center the predictions $\hat{\bm{y}}$ and subsequently drop the constant $c_2$. We will assume centered predictions in the following and use the correction $\hat{\bm{y}}^c = \mathcal{P}_{\bm{X}}^\bot \hat{\bm{y}}$.

\subsection{Details and Derivation of \cref{cor:constrOptim}} \label{app:theo1}

In the following, we describe how to derive the constrained optimization problem and subsequently our preferred way of solving it.

\paragraph{Deriving the constrained optimization problem}

To see that we can write problem 
\begin{equation}
    \hat{\bm{\gamma}}^c = \argmin_{\bm{\gamma}^c} \ell(\bm{y}, h(\bm{Z}\bm{\gamma}^c))~s.t.~\hat{\bm{\beta}}^c = \bm{0}
\end{equation}
with
\begin{equation} \label{eq:glm_copied}
    \hat{\bm{\beta}}^c = \argmin_{\bm{\beta}^c} \ell(\hat{\bm{y}}^c = h(\bm{Z}\hat{\bm{\gamma}}^c), h(\bm{X}\bm{\beta}^c))
\end{equation}
as
\begin{equation} \label{eq:constrOptim_copied}
  \begin{gathered}
    \hat{\bm{\gamma}}^c = \argmin_{\bm{\gamma}^c} \ell(\bm{y},h(\bm{Z\gamma^c}))\\ 
    s.t.~||(\bm{X}-n^{-1} \bm{1}_n\bm{1}_n^\top \bm{X})^\top h(\bm{Z\gamma^c})||_2^2 = 0,
\end{gathered}
\end{equation}
note that the statement in \cref{lemma:1} can be simplified by recognizing that \eqref{eq:betahat2} will also hold true if $\bm{X}^\top (\hat{\bm{y}}^c - \bm{\mu}) = \bm{X}^\top (\hat{\bm{y}}^c - c_2 \bm{1}_n) = \bm{0}$. The second term $\bm{X}^\top c_2 \bm{1}_n$ is always equal to zero if we center $\bm{X}$, i.e., use $(\bm{X}-n^{-1} \bm{1}_n\bm{1}_n^\top \bm{X})$. Hence, if $(\bm{X}-n^{-1} \bm{1}_n\bm{1}_n^\top \bm{X})^\top \hat{\bm{y}}^c = \bm{0}$ is fulfilled, \eqref{eq:betahat2} will also hold true. Now, instead of using $(\bm{X}-n^{-1} \bm{1}_n\bm{1}_n^\top \bm{X})^\top h(\bm{Z\gamma^c}) = \bm{0}$ as a constraint, introducing $p$ auxiliary variables when using the Lagrangian multiplier method (see details below), we can instead also use the constraint $||(\bm{X}-n^{-1} \bm{1}_n\bm{1}_n^\top \bm{X})^\top h(\bm{Z\gamma^c})||_2^2 = 0$, which implies the former, but only requires one Lagrangian multiplier variable.

\paragraph{Modified differential multiplier method (MDMM)}

Following \citet{platt1987constrained}, we first reformulate the constrained optimization problem \eqref{eq:constrOptim_copied} using Lagrangian multipliers, yielding
\begin{equation} \label{eq:lagrangian}
    \argmin_{\bm{\gamma},\lambda} \ell(\bm{y},h(\bm{Z\gamma})) + \lambda \underbrace{||(\bm{X}-n^{-1} \bm{1}_n\bm{1}_n^\top \bm{X})^\top h(\bm{Z\gamma})||_2^2}_{=:\mathcal{A}(\bm{\gamma})} =: \argmin_{\bm{\gamma},\lambda} \mathcal{E}(\bm{\gamma},\lambda).
\end{equation}
In order to optimize \eqref{eq:lagrangian}, we can use the basic differential multiplier method (BDMM) by performing gradient descent to optimize $\mathcal{E}$ w.r.t.~$\bm{\gamma}$ but gradient ascent for the auxiliary term $\lambda \mathcal{A}$ w.r.t.~$\lambda$, i.e., updating the parameters in each iteration $t\in\mathbb{N}$ as
\begin{equation} \label{eq:dmm}
    \begin{gathered}
        \bm{\gamma}^{[t]} = \bm{\gamma}^{[t-1]} - \nu^{[t]} \nabla_{\bm{\gamma}} \mathcal{E} (\bm{\gamma}^{[t-1]}; \lambda^{[t-1]}), \\
        \lambda^{[t]} = \lambda^{[t-1]} + \nu^{[t]} \mathcal{A}(\lambda^{[t-1]}; \bm{\gamma}^{[t-1]}),
    \end{gathered}
\end{equation}
with learning rate $\nu^{[t]}$. The \emph{modified} version MDMM of the BDMM suggested by \citet{platt1987constrained} combines the Lagrangian method with the penalty method augmenting the overall objective by a penalty term
\begin{equation}
    \mathcal{E}_P(\bm{\gamma},\lambda) := \mathcal{E}(\bm{\gamma},\lambda) + \frac{\zeta}{2} \mathcal{A}(\bm{\gamma})^2
\end{equation}
with damping factor $\zeta$ (that defaults to $1$), and thus adds another term $\zeta \mathcal{A}(\bm{\gamma})$ to the gradient descent direction in \eqref{eq:dmm}.

\subsection{Proof of Theorem~\ref{thm:piece}} \label{app:theo2}

Consider the ReLU activation function defined as $h(x) = x \cdot \mathds{1}(x \geq 0)$.
Using the fact that the MSE-optimal $\hat{\bm{\beta}}^c$ is found by
\begin{equation} \label{eq:pw}
    \hat{\bm{\beta}^c} = \argmin_{\bm{\beta}^c} \frac{1}{n} ||h(\bm{Z\gamma}) - h(\bm{X\beta}^c)||_2^2,
\end{equation}
we can first expand the objective which is minimized in \eqref{eq:pw} to
\begin{equation} \label{eq:pw2}
    \frac{1}{n} \left(h(\bm{Z\gamma})^\top h(\bm{Z\gamma}) - 2h(\bm{Z\gamma})^\top h(\bm{X\beta}^c) + h(\bm{X\beta}^c)^\top h(\bm{X\beta}^c)\right).
\end{equation}
The first dot product in \eqref{eq:pw2} is constant w.r.t.~$\bm{\beta}^c$ and the last dot product only determines the scaling of $\hat{\bm{\beta}}^c$, hence we focus on the mixed term. If, after feature correction $\bm{Z}^c = \mathcal{P}^\bot_{\bm{X}} \bm{Z}$, the mixed term 
\begin{equation} \label{eq:mixedzero}
    h(\bm{Z}^c\bm{\gamma})^\top h(\bm{X\beta}^c) = 0,
\end{equation}
the optimal $\hat{\bm{\beta}}^c$ in \eqref{eq:pw} will also be zero.

We can exploit properties of the ReLU function to show that this is indeed the case, using the fact that for any $\bm{a},\bm{b}\in\mathbb{R}^n$ it holds
\begin{equation}\label{eq:relu-trick}
\begin{split}
    \bm{a}^\top \bm{b} = \sum_i a_i b_i &= \sum_i \left[ h(a_i) h(b_i) + h(-a_i) h(b_i) + h(a_i) h(-b_i) + h(-a_i)h(-b_i) \right] \\
    &= \sum_i h(a_i) h(b_i) + \sum_i h(-a_i) h(b_i) + \sum_i h(a_i) h(-b_i) + \sum_i h(-a_i)h(-b_i) \\
    &= h(\bm{a})^\top h(\bm{b}) + h(-\bm{a})^\top h(\bm{b}) + h(\bm{a})^\top h(-\bm{b}) + h(-\bm{a})^\top h(-\bm{b}),
\end{split}
\end{equation}
which allows us to rewrite the left-hand side of \eqref{eq:mixedzero} as
\begin{equation}
\begin{split}
    & h(\bm{Z^c\gamma})^\top h(\bm{X\beta}^c) = (\bm{Z^c\gamma})^\top (\bm{X\beta}^c) - h(\bm{-Z^c\gamma})^\top h(\bm{X\beta}^c) - h(\bm{Z^c\gamma})^\top h(-\bm{X\beta}^c) - h(\bm{-Z^c\gamma})^\top h(-\bm{X\beta}^c)  \\
    \iff\,& h(\bm{Z^c\gamma})^\top h(\bm{X\beta}^c) = \underbrace{\bm{\gamma}^\top \bm{Z}^{\top} \mathcal{P}^\bot_{\bm{X}} \bm{X\beta}^c}_{=0} - h(\bm{-Z^c\gamma})^\top h(\bm{X\beta}^c) - h(\bm{Z^c\gamma})^\top h(-\bm{X\beta}^c) - h(\bm{-Z^c\gamma})^\top h(-\bm{X\beta}^c) \\
    \iff\,& h(\bm{Z^c\gamma})^\top h(\bm{X\beta}^c) = - h(\bm{-Z^c\gamma})^\top h(\bm{X\beta}^c) - h(\bm{Z^c\gamma})^\top h(-\bm{X\beta}^c) - h(\bm{-Z^c\gamma})^\top h(-\bm{X\beta}^c) \\
    \iff\,& h(\bm{Z^c\gamma})^\top h(\bm{X\beta}^c) + h(\bm{-Z^c\gamma})^\top h(\bm{X\beta}^c) + h(\bm{Z^c\gamma})^\top h(-\bm{X\beta}^c) + h(\bm{-Z^c\gamma})^\top h(-\bm{X\beta}^c)  = 0.
    \end{split}
\end{equation}
Since $h(\cdot) \geq 0$ for all inputs, the left side of the equation consists of a sum of products of non-negative terms, implying that all product terms must be equal to zero.

\subsection{Proof of Corollary~\ref{cor:tensor}} \label{app:tensor}

Using the $\text{vec}(\cdot)$ operation, we can first reformulate the evaluation model with uncorrected predictions $\text{vec}(\hat{\mathfrak{Y}}) \in \mathbb{R}^{n \cdot d}$, $ d = \prod_{r=1}^R d_r,$ as
\begin{equation} \label{eq:tosr3}
    \text{vec}(\hat{\mathfrak{Y}}) = \text{vec}(\bm{X} \times_1 \mathfrak{B}^c) = \text{vec}(\bm{X} \underbrace{\text{mat}(\mathfrak{B}^c)}_{p \times d}) = (\bm{I}_d \otimes \bm{X}) \text{vec}(\mathfrak{B}^c).
\end{equation}
The orthogonal projection matrix of $(\bm{I}_d \otimes \bm{X})$ is given by $(\bm{I}_{d} \otimes \mathcal{P}_{\bm{X}}^\bot)$. Premultiplying $\text{vec}(\hat{\mathfrak{Y}})$ by this term yields
\begin{equation}
(\bm{I}_{d} \otimes \mathcal{P}_{\bm{X}}^\bot)\text{vec}(\hat{\mathfrak{Y}}) = \text{vec}(\mathcal{P}_{\bm{X}}^\bot \underbrace{\text{mat}(\hat{\mathfrak{Y}})}_{n \times d}) = \text{vec}(\underbrace{\mathcal{P}_{\bm{X}}^\bot \times_1 \hat{\mathfrak{Y}}}_{:=\hat{\mathfrak{Y}}^c}).\end{equation}
The solution $\hat{\mathfrak{B}}^c$ to the evaluation model using corrected predictions $\hat{\mathfrak{Y}}^c$ can then be found by minimizing
\begin{equation}
    ||\text{vec}(\hat{\mathfrak{Y}}^c) - \text{vec}(\bm{X} \times_1 \mathfrak{B}^c)||_2^2,
\end{equation}
over $\mathfrak{B}^c$. The closed-form solution is given by
\begin{equation}
    \text{vec}(\hat{\mathfrak{B}}^c) = ((\bm{I}_d \otimes \bm{X})^\top (\bm{I}_d \otimes \bm{X}))^{-1} (\bm{I}_d \otimes \bm{X})^\top \text{vec}(\hat{\mathfrak{Y}}^c) = ((\bm{I}_d \otimes \bm{X})^\top (\bm{I}_d \otimes \bm{X}))^{-1} \underbrace{(\bm{I}_d \otimes \bm{X})^\top (\bm{I}_{d} \otimes \mathcal{P}_{\bm{X}}^\bot)}_{=\bm{0}}\text{vec}(\hat{\mathfrak{Y}}),
\end{equation}
proving that $\hat{\mathfrak{B}}^c \equiv \bm{0}$.

\subsection{Proof of \cref{cor:relu2}} \label{app:theo3}

Similar to the proof of \cref{thm:piece}, we start with the solution of the evaluation model given by
\begin{equation}
    \hat{\mathfrak{B}}^c = \argmin_{\mathfrak{B}^c} \frac{1}{n} ||h(\text{vec}(\bar{\mathfrak{Y}}^c)) - h(\text{vec}(\bm{X} \times_1 \mathfrak{B}^c))||_2^2,
\end{equation}
where $h(x) = x \cdot \mathds{1}(x \geq 0)$ denotes the ReLU function, and focus on the mixed term $h(\text{vec}(\bar{\mathfrak{Y}}^c))^\top h(\text{vec}(\bm{X} \times_1 \mathfrak{B}^c))$ when expanding the argument. Noting that this is again a dot product of vectors, we can use the same argument as before in \eqref{eq:relu-trick} for the proof of \cref{thm:piece}. In particular, the non-activated 
mixed term including the correction evaluates to
\begin{equation}
    \text{vec}(\mathcal{P}_{\bm{X}}^\bot \times_1 \bar{\mathfrak{Y}})^\top \text{vec}(\bm{X}\times_1 \mathfrak{B}^c) = (\text{vec} \bar{\mathfrak{Y}})^\top \underbrace{(\bm{I}_{d} \otimes \mathcal{P}_{\bm{X}}^\bot)(\bm{I}_d \otimes \bm{X})}_{\bm{0}} \, \text{vec}(\mathfrak{B}^c) = 0,
\end{equation}
and all other terms are again activated using the ReLU activation (and hence must be non-negative, but since their sum must be zero, all terms must be zero).


\section{Numerical Experiment Details} \label{app:model}

The code for reproducing results can be found on the first author's \href{https://github.com/davidruegamer/generalized_orthogonalization}{Github repository}.

\begin{table}[!ht]
    \centering
    \footnotesize
    \caption{Overview of all datasets used. The reference category refers to the category used as a reference in the $\mathcal{M}^e$ when encoding categorical variables.}
    \label{tab:datasets}
    \begin{tabular}{rccccc}
       & $n_{train}$  & $n_{test}$ & $p$ & $q$ & reference category \\ \hline
       Adult & 30162 & - & 5 & 45 & female, White\\
       Bank & 40195 & - & 1 & 49 & - \\
       Compas & 5855 & - & 6 & 13 & female, African-American\\
       Health Retirement & 38653 & - & 4 & 25 & female, married, Black \\
       MNISTred & 11872 & 1989 & 1 & 28 $\times$ 28 & red \\
       MIMIC & 181342 & 3041 & 4 & 111 & female, Asian \\
       UTKfaces & 16514 & 4129 & 5 & 256 & Asian \\
       Movie Reviews & 34167 & 8538 & 2 & 100 $\times$ 100 & female, female (main actor, movie director)
    \end{tabular}
\end{table}

\subsection{Adult Income Prediction}

For $\mathcal{M}^p$ we use the features age, employment status, education, marital status, relationship, and hours per week. Protected features are given in Table~\ref{tab:datasets}. All models are GLMs with canonical link function and binomial distribution, i.e., $\ell$ is the binary cross-entropy loss and $h()$ the sigmoid function.

\subsection{Fairness Benchmark}

For all methods, we set the unfairness amount to 0.05 (as 0 would result in an infinite penalty).

\subsection{Chest X-ray Embeddings}

We utilize the publicly available embeddings\footnote{https://doi.org/10.13026/pxc2-vx69} of the CXR foundation model proposed by \cite{sellergren2022simplified}.
The train and test partition is based on the recommended splits of the original MIMIC-CXR database.

Before using the embedding in a classifier, we apply singular value decomposition (SVD) to reduce the risk of overfitting in the 1376-dimensional embedding. We find that $q=111$ dimensions are the best trade-off between explained variance (98\%) and sparsity. The prediction model uses only these $q$ features while the correction function and evaluation model are defined by age, sex, and race.

\subsection{Face Recognition}

The embedding of the UTKFace dataset consists of the activations in the penultimate layer in a ResNet-50. 
Similar to the Chest X-ray application, we first reduce the 2048-dimensional embedding to $q=32$ features using an SVD. These $q$ features explain 95\% of the embedding's variance. 
The train and test split is based on a random 80/20 partitioning.
We use these embedding features in the prediction model while using age and race for the correction function and evaluation model.

\subsection{Movie Reviews}

We first extract the first actor's sex and the director's sex from the provided metadata. We then tokenize the movie descriptions using 1000 words and an embedding size of 100 with a maximum length of 100 words for each description. Padding is done at the end of sentences with less than 100 words. We further remove stop words and punctuations from sentences and transform all texts to lowercase. We use the vote count as a Poisson-distributed outcome and partition the data into 80/20\% for training and testing. 

The LSTM model is defined by an embedding layer with embedding size 100, an LSTM layer with 50 units and ReLU activation, a dropout layer with 0.1 dropout rate, a dense layer with 25 units and ReLU activation, a dropout layer with 0.2 dropout rate, a dense layer with 5 units and ReLU activation, a dropout layer with 0.3 dropout rate, and a final dense layer with 1 unit and exponential activation. The network is trained for a maximum of 1000 epochs with early stopping using Adam with a learning rate of 1e-6, a batch size of 128, and Poisson loss. 

For the orthogonalization, we use the same architecture but insert an orthogonalization operation between the embedding layer and the LSTM layer. As features $\bm{Z}$ we use only the movie descriptions, whereas the protected features $\bm{X}$ consist of the gender of the protagonist and director.

\subsection{Online Orthogonalization}

We subsample the MNIST dataset to observations with labels 0 or 9. Based on the coloring described in the main part of the paper, we train a CNN defined by a Conv2D layer with 8 filters, kernel size of $3\times 3$, and ReLU activation, followed by the orthogonalization operation for the network that uses a correction. The output (with or without correction) is then fed into another Conv2D layer with 16 filters, kernel size of $3\times 3$, and ReLU activation, followed by Max pooling of size $2 \times 2$ and a flattening operation. Finally, the classification head comprises a dense layer with 64 units and ReLU activation followed by the output layer with one unit and sigmoid activation.

The prediction model only uses the images and, if corrected, also the color information to perform the orthogonal projection for training. The networks are trained for 100 epochs with early stopping 
and a batch size of 128. Early stopping is based on a $20\%$ validation split and a patience of 25. The evaluation model is optimized using quadratic programming and only uses the color red information.

\subsection{Runtimes}

Table~\ref{tab:runtime} compares the runtimes of different methods on the fairness benchmark dataset.

\begin{table}[htbp]
  \centering
  \caption{Runtimes on the fairness benchmark dataset (in seconds).}
    \begin{tabular}{lcccc}
    \toprule
    Method & adult & bank & compas & health.retirement \\
    \midrule
    Komiyama1 & 3.62 & 2.37 & 0.65 & 1.00 \\
    Komiyama2 & 5.66 & 2.68 & 1.19 & 14.17 \\
    Berk    & 1.13 & 2.59 & 1.18 & 3.52 \\
    Zafar   & 15.02 & 0.56 & 12.02 & - \\
    Ours    & 1.11 & 13.20 & 0.48 & 3.84 \\
    \bottomrule
    \end{tabular}%
  \label{tab:runtime}%
\end{table}%

For the colorized MNIST dataset, the average fitting time until early stopping (with standard deviation in brackets) of both networks was 1.08 (0.528) minutes with orthogonalization and 0.40 (0.006) minutes without orthogonalization. We can see that the orthogonalization takes on average longer to ``converge''. This can mainly be attributed to the fact that it runs for more iterations and not the additional operations required to perform the orthogonalization.

\section{Additional Results}

\subsection{Synthetic Data} \label{app:syn}

First, we simulate data from a Bernoulli and Poisson distribution to check whether our approach provides the proposed protection for features $\bm{X}$ using a canonical activation function $h$ and either $\mathcal{C}^l$ or $\mathcal{C}^h$. We generate features $\bm{X}$ from a standard normal distribution and generate features $\bm{Z}$ correlated to $\bm{X}$ using the transformation $\bm{X} = \rho \bm{Z}_{[:,1:q]} + \bm{E}$, where $\rho \in \{0,1,2\}$ and $\bm{E}$ are realizations of a standard normal distribution with $\dim(\bm{E}) = \dim(\bm{Z})$. We examine different settings for $p\in\{5,10\}$, $q\in\{10,100\}$, and $n\in\{200, 1000, 5000, 10000\}$, and repeat every setting 10 times with different random seed.

\begin{figure}[!ht]
    \centering
    \includegraphics[width=\columnwidth]{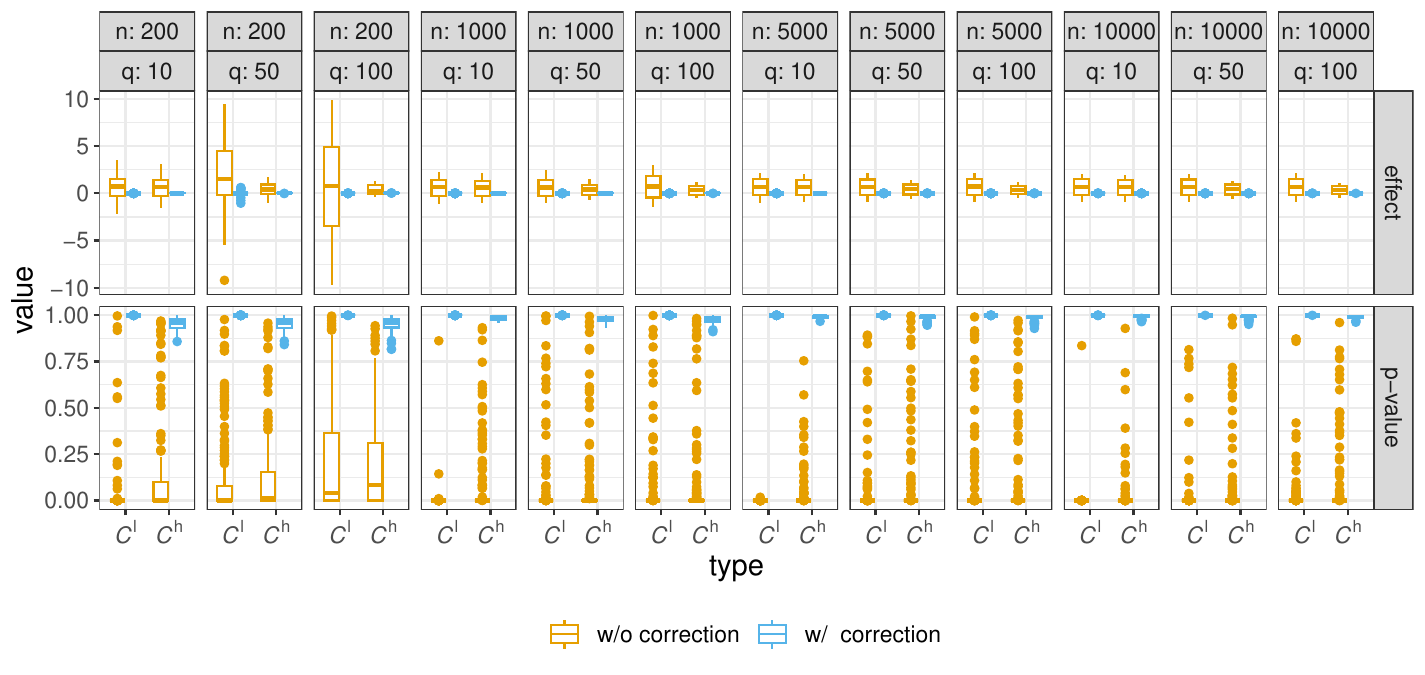}
    \caption{Distribution of estimated coefficients (top row) and corresponding p-values (bottom row) for different data sizes $n$ and features $q$ (columns) for the binary classification task (Bernoulli distribution). Boxplots summarize all different values of $p$ and the 10 simulation repetitions.}
    \label{fig:glmsim}
\end{figure}

\paragraph{Results}
Figure~\ref{fig:glmsim} shows the results of the simulation for the classification task and $\rho=2$ (the count response task in the Appendix is qualitatively the same) with and without correction for the model's pre-activation (corrected with $\mathcal{C}^l$) and post-activation values (corrected with $\mathcal{C}^h$). Results confirm that not only are coefficients estimated to be close to zero after correction, but their p-value is also close to 1 in most cases, suggesting no significant influence of protected features in the predicted value $\hat{\bm{y}}^c$. In contrast, without correction, coefficients tend to be non-zero and ``highly significant'' (with a very small p-value).

\begin{figure}[!ht]
    \centering
    \includegraphics[width=0.95\columnwidth]{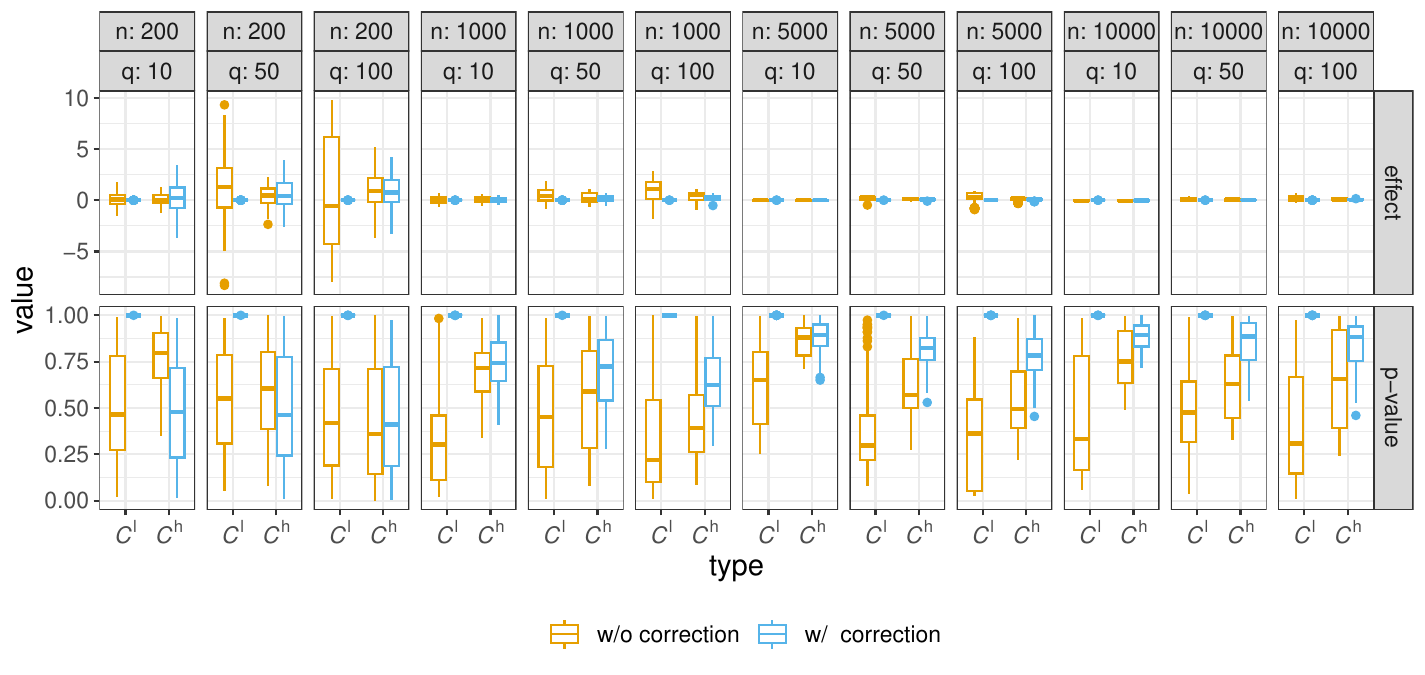}
    \caption{Distribution of estimated coefficients (top row) and corresponding p-values (bottom row) for different data sizes $n$ (columns) for the classification task and $\rho = 0$. Boxplots summarize all different values of $p$, $q$, and the 10 simulation repetitions.}
    \label{fig:glmsim2}
\end{figure}

\begin{figure}[!ht]
    \centering
    \includegraphics[width=0.95\columnwidth]{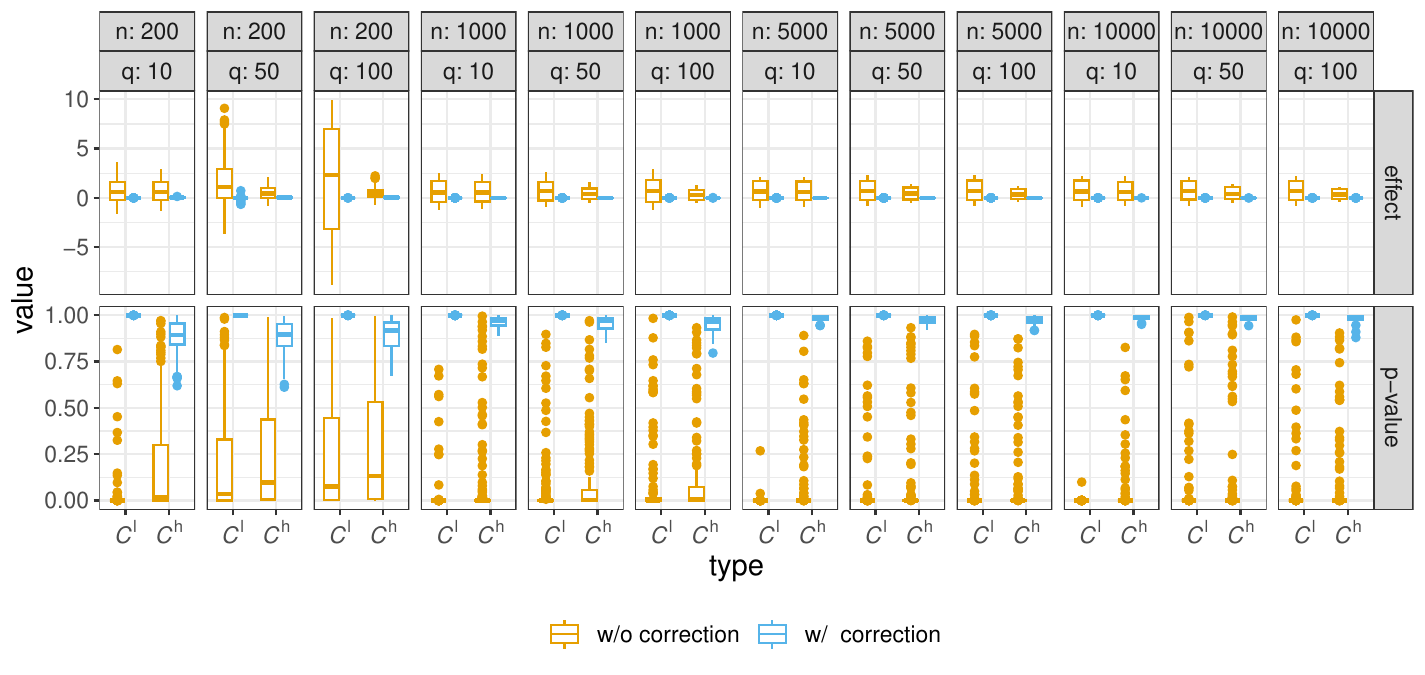}
    \caption{Distribution of estimated coefficients (top row) and corresponding p-values (bottom row) for different data sizes $n$ (columns) for the classification task and $\rho = 0$. Boxplots summarize all different values of $p$ and the 10 simulation repetitions.}
    \label{fig:glmsim3}
\end{figure}

\begin{figure}[!ht]
    \centering
    \includegraphics[width=0.95\columnwidth]{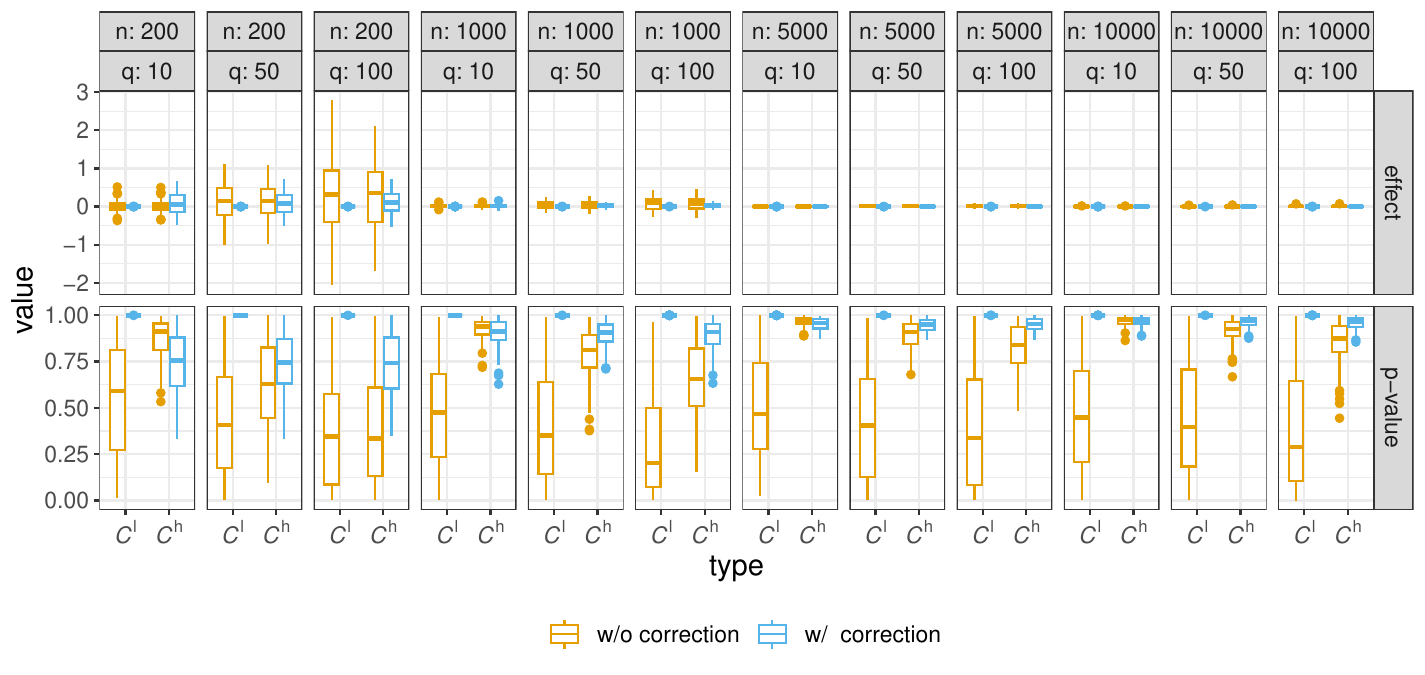}
    \caption{Distribution of estimated coefficients (top row) and corresponding p-values (bottom row) for different data sizes $n$ (columns) for the count task (Poisson) and $\rho = 0$. Boxplots summarize all different values of $p$ and the 10 simulation repetitions.}
    \label{fig:glmsim4}
\end{figure}

\begin{figure}[!ht]
    \centering
    \includegraphics[width=0.95\columnwidth]{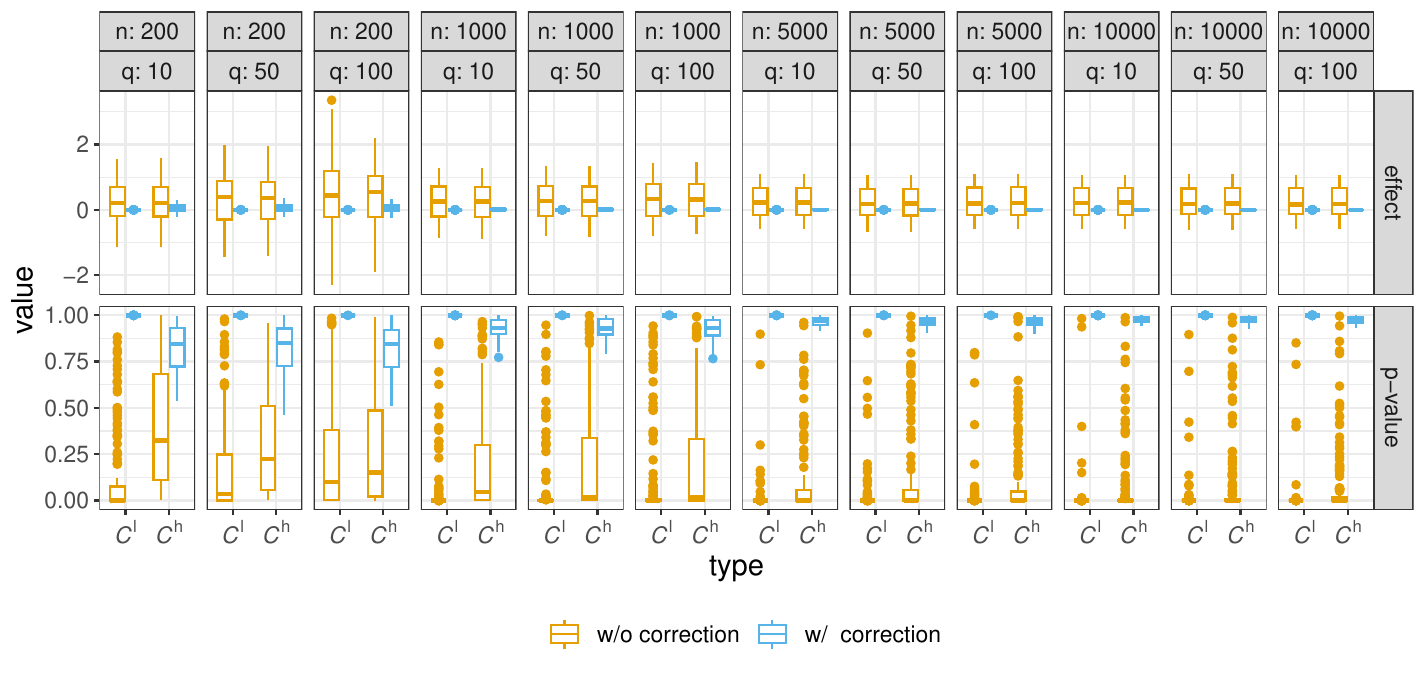}
    \caption{Distribution of estimated coefficients (top row) and corresponding p-values (bottom row) for different data sizes $n$ (columns) for the count task (Poisson) and $\rho = 1$. Boxplots summarize all different values of $p$ and the 10 simulation repetitions.}
    \label{fig:glmsim5}
\end{figure}

\begin{figure}[!ht]
    \centering
    \includegraphics[width=0.95\columnwidth]{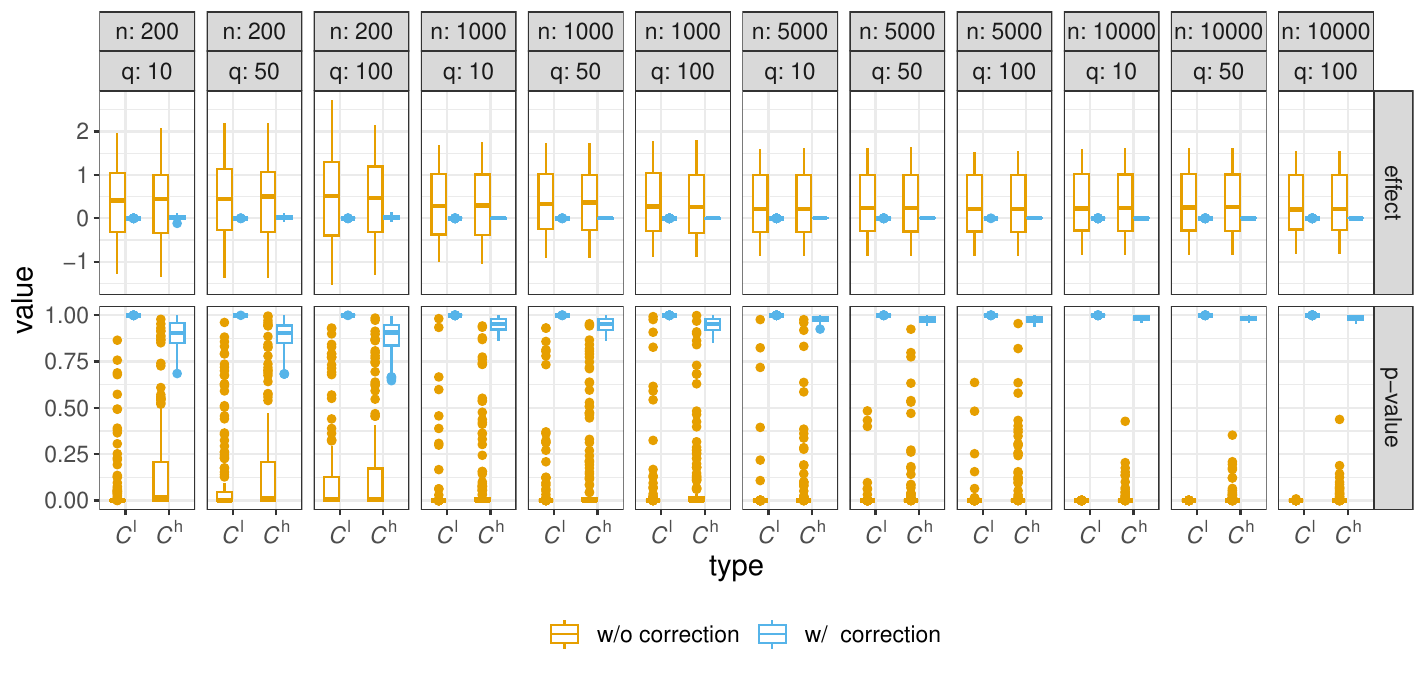}
    \caption{Distribution of estimated coefficients (top row) and corresponding p-values (bottom row) for different data sizes $n$ (columns) for the count task (Poisson) and $\rho = 2$. Boxplots summarize all different values of $p$ and the 10 simulation repetitions.}
    \label{fig:glmsim6}
\end{figure}

Similar results are obtained when using $\rho=0$ or $\rho=1$ for a logistic regression model (cf.~Figure~\ref{fig:glmsim2} and~\ref{fig:glmsim3}, respectively) and also for $\rho\in\{0,1,2\}$ for a count regression (Figures~\ref{fig:glmsim4}-\ref{fig:glmsim6}).

\clearpage

\end{document}